\documentclass{article}

\usepackage{PRIMEarxiv}

\usepackage[utf8]{inputenc} % allow utf-8 input
\usepackage[T1]{fontenc}    % use 8-bit T1 fonts
\usepackage{hyperref}       % hyperlinks
\usepackage{url}            % simple URL typesetting
\usepackage{booktabs}       % professional-quality tables
\usepackage{amsfonts}       % blackboard math symbols
\usepackage{nicefrac}       % compact symbols for 1/2, etc.
\usepackage{microtype}      % microtypography
\usepackage{lipsum}
\usepackage{fancyhdr}       % header
\usepackage{graphicx}       % graphics
\graphicspath{{media/}}     % organize your images and other figures under media/ folder

\usepackage{color}
\usepackage{mdframed} 
\usepackage[usenames,dvipsnames]{xcolor}
\usepackage{amssymb}   
\usepackage{todonotes}
\usepackage{titlesec}
\usepackage{glossaries}
\usepackage{algorithm}
\usepackage{algpseudocode}
\usepackage{tabularx,multirow}
\usepackage{times}
\usepackage{epsfig}
\usepackage{nicefrac}
\usepackage{adjustbox} %
\newcolumntype{R}[2]{%
    >{\adjustbox{angle=#1,lap=\width-(#2)}\bgroup}%
    l%
    <{\egroup}%
}
\newcommand*\rot{\multicolumn{1}{R{90}{1em}}}

\title{Geometric Constraints in Deep Learning Frameworks: A Survey
}

\author{
  Vibhas K Vats, David J Crandall \\
  Indiana University Bloomington \\
  \texttt{\{vkvats, djcran\}@iu.edu} \\
}

\begin{document}
\maketitle

%%%%%%%%%%%%%%%%%%%%%%%%%%%%%%%%% Abstract %%%%%%%%%%%%%%%%%%%%%%%%%%%%%%%%

\begin{abstract}
Stereophotogrammetry \cite{photogra_wiki} is an established technique for scene understanding. 
Its origins go back to at least the 1800s when people first started to investigate 
using photographs to measure the physical properties of the world. 
Since then, thousands of approaches have been explored. 
The classic geometric technique of Shape from Stereo is 
built on using geometry to define constraints on scene and camera
deep learning without any attempt to explicitly model the geometry. 
In this survey, we explore geometry-inspired deep learning-based frameworks. 
We compare and contrast geometry enforcing constraints integrated 
into deep learning frameworks for depth estimation and other closely 
related vision tasks. We present a new taxonomy for prevalent 
geometry enforcing constraints used in modern deep learning frameworks. 
We also present insightful observations and potential future research directions.
\end{abstract}

%%%%%%%%%%%%%%%%%%%%%%%%%%%%%%%%% Key words %%%%%%%%%%%%%%%%%%%%%%%%%%%%%%%%
% keywords can be removed

%%%%%%%%%%%%%%%%%%%%%%%%%%%%%%%%% Main content %%%%%%%%%%%%%%%%%%%%%%%%%%%%%%%%

\section{Introduction}
\label{sec:intro}
Traditional stereo or multi-view stereo (MVS) depth estimation 
 methods rely on solving for 
photometric and geometric consistency constraints across view(s) 
for consistent depth estimation 
\cite{Furukawa2010AccurateDA, Galliani2015fusibile, tola2011largescale, Johannes2016pixelwise, boykov2001fast, hartley2003multiple, luong2001geometry, szeliski2022computer}. 
With the phenomenal rise of deep learning frameworks \cite{lecun2015deep} 
like Convolutional Neural Networks (CNNs) \cite{lecun1995convolutional}, 
Recurrent Neural Networks (RNNs) \cite{malhotra2015long}, and 
Vision Transformers (ViTs) \cite{dosovitskiy2020image}, 
which can extract deep local and high-level features, 
the requirement to apply photometric and geometric consistency constraints 
has significantly reduced, especially in supervised depth estimation methods. 
Compared to traditional feature extraction methods, deep learning-based features significantly improve feature matching, leading to significant improvements in depth estimates 
\cite{yao2018mvsnet,gu2020cascade,ding2022transmvsnet,kendall2017end,dong2024ppea,dong2024mal}. 
However, the application of geometric constraints in these deep learning-based frameworks remains limited to 
the use of plane-sweep algorithm \cite{collins1996space} for a majority 
of supervised stereo and MVS methods.

\begin{figure*}[t]
    \centering
    \includegraphics[width=0.8\textwidth]{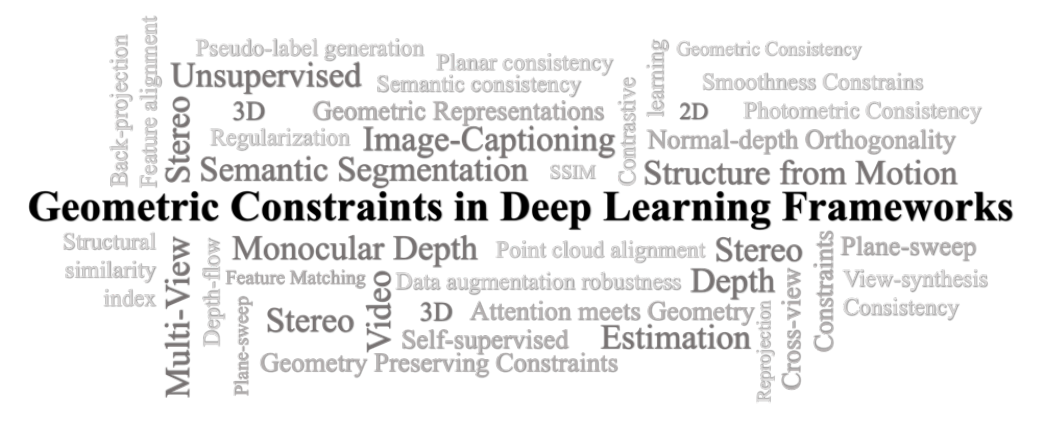}
    \caption{Overview of geometry-inspired deep learning-based constraints surveyed in this paper.
    Specifically, we discuss geometry-inspired constraints applied in deep learning frameworks for 
    depth estimation problems. We also discuss concepts that are used in closely related vision tasks
    like Structure from Motion, Semantic Segmentation, 
    Video depth estimation, Image Captioning, etc. This word cloud shows all geometry-inspired 
    concepts covered in this survey.}
    \label{fig:overview-word-collage}
% \end{figure*}

% \begin{figure*}[t]
\begin{center}
    \includegraphics[width=1\textwidth]{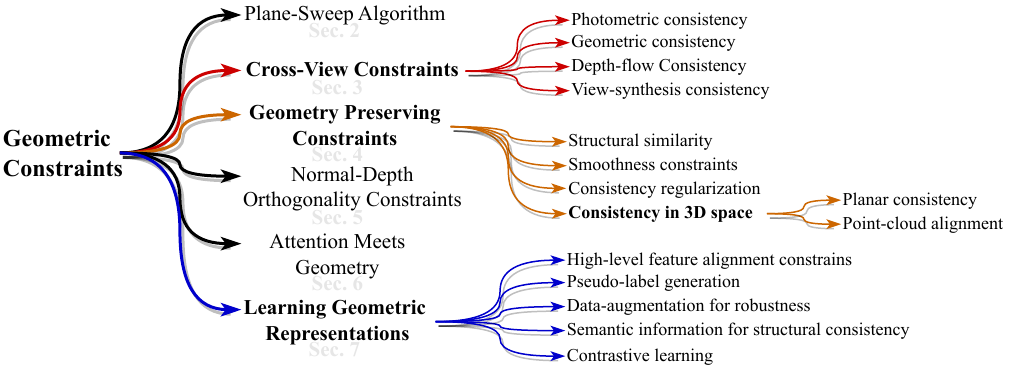}
    \caption{Our taxonomy of geometry-inspired constraints used in 
    deep learning-based depth estimation and other closely related vision tasks.}
    \label{fig:taxonomy}
\end{center}
\end{figure*}

In a typical supervised stereo or MVS depth estimation framework, 
the plane-sweep algorithm is applied to create a matching (cost) volume, 
which is then aggregated based on a metric. The aggregated volume (i.e. the cost 
volume) is then regularized using 3D-CNNs or RNNs to produce a coherent estimate.
The lack of ground truth in unsupervised/self-supervised depth estimation methods do not 
allow such freedom. Photometric and geometric consistency constraints 
remain a key part of unsupervised frameworks 
\cite{xu2021self, garg2016unsupervised, yang2017unsupervised, yang2021self, huang2021m3vsnet, dong2024ppea, dong2024mal}.
Some  other closely related problems, like structure from motion \cite{chen2019self, cheng2017segflow}, 
video-depth estimation \cite{mahjourian2018unsupervised, godard2017unsupervised}, 
semantic-segmentation \cite{eigen2015predicting, dovesi2020real, yang2018segstereo, xu2021self}, and
monocular depth estimation \cite{naderi2022monocular, bauer2021nvs, dong2024ppea}, also apply 
various geometric constraints for a consistent result. In this survey, 
we focus on all such methods that integrate photometric or geometric 
constraints in deep learning-based frameworks and 
are closely related to the depth estimation problem. Fig. \ref{fig:overview-word-collage} 
shows the collection of such geometric constraints and their associated problems that 
are covered in this survey. We discuss the theory and mathematical formulation of all
these concepts and present a carefully crafted taxonomy, as shown in Fig. \ref{fig:taxonomy}.

\begin{table}[t]
  \begin{center}
    {\footnotesize{
\begin{tabular}{llcccc}
\toprule
Constraints & Sub-Constraints & Pre-Process. & Post-Process. & Obj. Fn  & In-Network \\
\midrule
Plane-Sweep Algo. & -- & -- & -- & -- & $\checkmark$\\
\cmidrule{2-6}
    & Photometric Constraints & $\checkmark$ & $\checkmark$ & $\checkmark$ & --\\
\multirow{1}{*}{Cross-View} & Geometric Constraints & $\checkmark$ & $\checkmark$ & $\checkmark$ & --\\
\multirow{1}{*}{Constraints} & Depth-to-Flow Constraints & -- & -- & $\checkmark$ & --\\
    & RGB-to-Flow Constraints & -- & -- & $\checkmark$ & --\\
    & View Synthesis Constraints & -- & $\checkmark$ & $\checkmark$ & $\checkmark$\\
\cmidrule{2-6}
    & Structural Similarity Index & -- & -- & $\checkmark$ & $\checkmark$\\
    & Edge-Aware Smoothness & -- & -- & $\checkmark$ & --\\
\multirow{1}{*}{Geometry Preserving} & Consistency Regularization & -- & -- & $\checkmark$ & --\\
\multirow{1}{*}{Constraints}  & Planar Consistency & $\checkmark$ & $\checkmark$ & $\checkmark$ & --\\
    & Point Cloud Alignment & -- & -- & $\checkmark$ & $\checkmark$\\
\cmidrule{2-6}
\multirow{1}{*}{Normal-Depth} & Depth-to-Normal &$\checkmark$  & $\checkmark$ & $\checkmark$ & --\\
\multirow{1}{*}{Orthogonality Const.}  & Normal-to-Depth & $\checkmark$ & $\checkmark$ & $\checkmark$ & --\\
    & Normal-Depth Joint Opt. & $\checkmark$ & $\checkmark$ & $\checkmark$ & --\\
\cmidrule{2-6}
Attention \& Geometry & -- & -- & -- & -- & $\checkmark$\\
\cmidrule{2-6}
 & High-Level Feature Alignment & -- & -- & $\checkmark$ & $\checkmark$\\
    & Pseudo-Label Cross-View & -- & -- & $\checkmark$ & $\checkmark$\\
\multirow{1}{*}{Learning Geometric} & Data-Augmented Robustness & $\checkmark$ & -- & $\checkmark$ & --\\
\multirow{1}{*}{Representations}  & Semantic-Guided Structure & -- & -- & $\checkmark$ & $\checkmark$\\
    & Geometric Representation & -- & -- & $\checkmark$ & --\\
\bottomrule
\end{tabular}
}}

\caption{Our taxonomy organizes geometry-inspired constraints into six main constraints, some of which are further divided into sub-constraints. 
This table highlights the most common ways these sub-constraints 
are incorporated into deep learning frameworks. 
Specifically, we classify the sub-constraints into four primary use cases: 
as a pre-processing step, as a post-processing step, as part of the objective function, 
or as an integrated component within a deep neural network. This categorization provides 
a clear understanding of where and how these constraints are typically applied to enhance the learning process in geometry-aware models}
\label{table:organis-and-fit}
\end{center}
\end{table}

We employ `\textit{backward citation tracking},' also known as `\textit{citation chaining},' 
for each geometry-based constraint discussed in this survey. 
Specifically, we begin with recent papers and trace their references 
to identify the original work that introduced the concepts. 
We focus on selecting relevant papers published in prominent conferences and journals, 
excluding preprints that are only available on ArXiv. 
Our curated list spans two distinct eras: the \textit{early era}, 
which includes seminal papers that first introduced these concepts, 
and the \textit{modern era}, which consists of recent works (from the past 6–8 years) 
that apply these concepts, either individually or in combination with other 
geometry-based ideas, within learning-based frameworks. 
To identify relevant papers, we use \textit{Google Scholar} 
with keywords such as `\textit{geometric constraints}' 
combined with terms like `\textit{learning-based frameworks},' 
`\textit{depth estimation},' and `\textit{multi-view stereo},' 
ensuring a comprehensive and up-to-date review of the field.

Throughout this paper, we focus on geometry-enforcing concepts 
used across different problems that either do depth estimation or are closely 
related to depth estimation problems. 
We only discuss the specific concepts used 
and their relevance to stereo or MVS depth estimation frameworks. 
We also specify efficacy 
in handling cluttered backgrounds, repetitive patterns, and texture-less regions 
for each geometric constraint that can be used as an objective function. 
We provide two comprehensive snapshots of the surveyed methods 
in Tables \ref{table:organis-and-fit} and \ref{table:methods-and-concept}. 
Table \ref{table:organis-and-fit} offers a detailed view of how geometry-based 
constraints can be integrated into learning-based frameworks. 
These constraints are categorized into four primary applications: 
as pre-processing steps, post-processing steps, objective functions, 
or integral components within a network. Each sub-constraint discussed 
in this survey is classified into one or more categories.
Table \ref{table:methods-and-concept} summarizes key methods, 
showcasing the specific types of geometry-based constraints they incorporate. 
Notably, many methods employ multiple sub-constraints, 
often spanning different categories. 
Furthermore, some geometry-based 
constraints are adapted from unrelated domains, 
such as image captioning, structure from motion, 
and semantic segmentation, and have only recently been explored 
in the context of depth estimation. 
This cross-domain borrowing creates further 
inconsistencies, making it difficult to provide 
a meaningful quantitative comparison specific to depth estimation tasks.

%%%%%%%%%%%%%%%%%%%%%%%%%%%%%%%%%%%%%%%%%%%%%%%%%%%%%%%%%%%%%%%%%%%%%%%%%%%%%%%%%%%%%%%%%%%%%%%%%%%%%%%%%%%%%%%%%%%%%%%%%

\begin{table}[p]
  \begin{center}
    {\tiny{
\begin{tabular}{lccccccccccccccccccc}
\toprule
\multirow{3}{*}{\bf{Methods}} & & \multicolumn{4}{c}{\bf{Cross-View}} & \multicolumn{5}{c}{\bf{Geometry Preserving}} & \multicolumn{3}{c}{\bf{Normal-Depth}} & & \multicolumn{5}{c}{\bf{Learning Geometric}}\\
 & & \multicolumn{4}{c}{\bf{Constraints}} & \multicolumn{5}{c}{\bf{Constraints}} & \multicolumn{3}{c}{\bf{Orthogonal}} & & \multicolumn{5}{c}{\bf{Representations}}\\
\cmidrule{3-14}
\cmidrule{16-20}
& \rot{\bf{Plane-Sweep Algorithm}} & \rot{Photometric Constraints} & \rot{Geometric Constraints} & \rot{Depth-Flow Consistency} & \rot{View Synthesis Constraints} & \rot{Structural Similarity Index} & \rot{Edge-Aware Smoothness} & \rot{Consistency Regularization} & \rot{Planar Consistency} & \rot{Point-Cloud Alignment} & \rot{Depth-to-Normal} & \rot{Normal-to-Depth} & \rot{Normal-Depth Joint Opt.} & \rot{\bf{Attention \& Geometry}} & \rot{High-Level Feature Align.} & \rot{Pseudo-Label Cross View} & \rot{Data-Augmentation} & \rot{Semantic-Guided Structure} & \rot{Geometric Representation} \\
\midrule
NVS-MonoDepth~\cite{bauer2021nvs} & -- & -- & -- & -- & $\checkmark$ & -- & -- & -- & -- & -- & -- & -- & -- & -- & -- & -- & -- & -- & --\\
Besl and McKay~\cite{besl1992method} & -- & -- & -- & -- & -- & -- & -- & -- & -- & $\checkmark$ & -- & -- & -- & -- & -- & -- & -- & -- & --\\
Brox et al.~\cite{brox2004high}& -- & -- & -- & -- & -- & -- & $\checkmark$ & -- & -- & -- & -- & -- & -- & -- & -- & -- & -- & -- & --\\
Chen et al.~\cite{chen2019point} & -- & -- & -- & $\checkmark$ & -- & -- & -- & -- & -- & -- & -- & -- & -- & -- & -- & -- & -- & -- & --\\
Chen et al.~\cite{chen2021fixing}  & -- & -- & $\checkmark$ & -- & -- & -- & -- & -- & -- & -- & -- & -- & -- & -- & -- & -- & -- & -- & --\\
Chen and Medioni~\cite{chen1991object} & -- & -- & -- & -- & -- & -- & -- & -- & -- & $\checkmark$ & -- & -- & -- & -- & -- & -- & -- & -- & --\\
Chen et al.~\cite{chen2019self}  & -- & -- & $\checkmark$ & -- & -- & $\checkmark$ & -- & -- & -- & -- & -- & -- & -- & -- & -- & -- & -- & -- & --\\
Segflow\cite{cheng2017segflow} & -- & -- & -- & -- & -- & -- & -- & -- & -- & -- & -- & -- & -- & -- & -- & -- & -- & $\checkmark$ & --\\
MVS$^2$~\cite{dai2019mvs2} & $\checkmark$ & $\checkmark$ & -- & -- & $\checkmark$ & -- & $\checkmark$ & -- & -- & -- & -- & -- & -- & -- & -- & -- & -- & -- & --\\
TransMVSNet~\cite{ding2022transmvsnet} & $\checkmark$ & -- & -- & -- & -- & -- & -- & -- & -- & -- & -- & -- & -- & -- & -- & -- & -- & -- & --\\
PatchMVSNet~\cite{dong2022patchmvsnet} & $\checkmark$ & $\checkmark$ & -- & -- & -- & -- & $\checkmark$ & -- & -- & -- & -- & -- & -- & -- & -- & -- & -- & -- & --\\
Gong et al.~\cite{dong2022geometry} & -- & -- & -- & -- & -- & $\checkmark$ & -- & -- & -- & -- & -- & -- & -- & -- & -- & -- & -- & -- & --\\
Dovesi et al.~\cite{dovesi2020real} & -- & -- & -- & -- & -- & -- & -- & -- & -- & -- & -- & -- & -- & -- & -- & -- & -- & $\checkmark$ & --\\
Eigen and Fergus~\cite{eigen2015predicting}& -- & -- & -- & -- & -- & -- & $\checkmark$ & -- & -- & -- & -- & $\checkmark$ & -- & -- & -- & -- & -- & -- & --\\
Engel et al.~\cite{engel2018DSO}  & -- & -- & -- & -- & -- & -- & -- & -- & $\checkmark$ & -- & -- & -- & -- & -- & -- & -- & -- & -- & --\\
Fan et al.~\cite{fan2023contrastive}  & -- & -- & -- & -- & -- & -- & -- & -- & -- & -- & -- & -- & -- & -- & -- & -- & -- & -- & $\checkmark$\\
Felzenszwalb and Huttenlocher~\cite{felzenszwalb2004efficient}  & -- & -- & -- & -- & -- & -- & -- & -- & $\checkmark$ & -- & -- & -- & -- & -- & -- & -- & -- & -- & --\\
Fouhey et al.~\cite{fouhey2013data}& -- & -- & -- & -- & -- & -- & -- & -- & -- & -- & $\checkmark$ & -- & $\checkmark$ & -- & -- & -- & -- & -- & --\\
Gallup et al.~\cite{gallup2007real} & $\checkmark$ & -- & -- & -- & -- & -- & -- & -- & -- & -- & -- & -- & -- & -- & -- & -- & -- & -- & --\\
Garg et al.~\cite{garg2016unsupervised}& -- & -- & -- & -- & -- & -- & $\checkmark$ & $\checkmark$ & -- & -- & -- & -- & -- & -- & -- & -- & $\checkmark$ & -- & --\\
Godard et al.~\cite{godard2017unsupervised}& -- & -- & -- & -- & -- & -- & $\checkmark$ & -- & -- & -- & -- & -- & -- & -- & -- & -- & -- & -- & --\\
Godard et al.~\cite{godard2019digging} & -- & -- & -- & $\checkmark$ & -- & -- & -- & -- & -- & -- & -- & -- & -- & -- & -- & -- & -- & -- & --\\
CasMVSNet~\cite{gu2020cascade}& $\checkmark$ & -- & -- & -- & -- & -- & -- & -- & -- & -- & -- & -- & -- & -- & -- & -- & -- & -- & --\\
Guo et al.~\cite{guo2020normalized}& -- & -- & -- & -- & -- & -- & -- & -- & -- & -- & -- & -- & -- & $\checkmark$ & -- & -- & -- & -- & --\\
M$^3$VSNet~\cite{huang2021m3vsnet} & $\checkmark$ & $\checkmark$ & -- & -- & -- & -- & $\checkmark$ & -- & -- & -- & -- & $\checkmark$ & -- & -- & $\checkmark$ & -- & -- & -- & --\\
Kendall et al.~\cite{kendall2017end} & $\checkmark$ & -- & -- & -- & -- & -- & -- & -- & -- & -- & -- & -- & -- & -- & -- & -- & -- & -- & --\\
Khot et al.~\cite{khot2019learning} & -- & $\checkmark$ & -- & -- & -- & $\checkmark$ & -- & -- & -- & -- & -- & -- & -- & -- & -- & -- & -- & -- & --\\
Kusupati et al.~\cite{kusupati2020normal} & -- & -- & -- & -- & -- & ----& -- & -- & -- & -- & -- & -- & $\checkmark$ & -- & -- & -- & -- & -- & --\\
Lee et al.~\cite{lee2021attentive}  & -- & -- & -- & -- & -- & -- & -- & -- & -- & -- & -- & -- & -- & -- & -- & -- & -- & -- & $\checkmark$\\
DS-MVSNet~\cite{li2022ds} & $\checkmark$ & $\checkmark$ & -- & -- & $\checkmark$ & $\checkmark$ & -- & -- & -- & -- & -- & -- & -- & -- & -- & -- & -- & -- & --\\
Liu et al.~\cite{liu2023geometric}  & -- & $\checkmark$ & -- & -- & -- & -- & -- & -- & -- & -- & -- & -- & -- & -- & -- & $\checkmark$ & -- & -- & --\\
Liu et al.~\cite{liu2020learning} & -- & -- & -- & $\checkmark$ & -- & -- & -- & -- & -- & -- & -- & -- & -- & -- & -- & -- & -- & -- & --\\
Mahjourian et al.~\cite{mahjourian2018unsupervised} & -- & -- & -- & -- & -- & $\checkmark$ & $\checkmark$ & -- & -- & $\checkmark$ & -- & -- & -- & -- & -- & -- & -- & -- & --\\
Loung and Faugeras~\cite{luong2001geometry} & $\checkmark$ & -- & -- & -- & -- & -- & -- & -- & -- & -- & -- & -- & -- & -- & -- & -- & -- & -- & --\\
Mallick et al.~\cite{mallick2020learning}  & -- & $\checkmark$ & -- & -- & -- & $\checkmark$ & -- & -- & -- & -- & -- & -- & -- & -- & -- & -- & -- & -- & --\\
Naderi et al.~\cite{naderi2022monocular}& -- & -- & -- & -- & -- & -- & -- & -- & -- & -- & -- & $\checkmark$ & -- & $\checkmark$ & -- & -- & -- & -- & --\\
GeoNet~\cite{qi2018geonet}& -- & -- & -- & -- & -- & -- & -- & -- & -- & -- & $\checkmark$ & $\checkmark$ & $\checkmark$ & -- & -- & -- & -- & -- & --\\
Ruhkamp et al.~\cite{ruhkamp2021attention}& -- & -- & -- & -- & -- & -- & -- & -- & -- & -- & -- & -- & -- & $\checkmark$ & -- & -- & -- & -- & --\\
Rusinkiewicz and Levoy~\cite{Rusinkiewicz2001efficient} & -- & -- & -- & -- & -- & -- & -- & -- & -- & $\checkmark$ & -- & -- & -- & -- & -- & -- & -- & -- & --\\
Saito and Kanade~\cite{saito1999shape} & $\checkmark$ & -- & -- & -- & -- & -- & -- & -- & -- & -- & -- & -- & -- & -- & -- & -- & -- & -- & --\\
Shim and Kim~\cite{shim2021learning} & -- & -- & -- & -- & -- & -- & -- & -- & -- & -- & -- & -- & -- & -- & -- & -- & -- & -- & $\checkmark$\\
Silberman et al.~\cite{silberman2012indoor}& -- & -- & -- & -- & -- & -- & -- & -- & -- & -- & -- & $\checkmark$ & -- & -- & -- & -- & -- & -- & --\\
LoFTR~\cite{sun2021loftr}& --& --& --& --& --& --& --& --& --& --& --& --& --& $\checkmark$ & --& --& --& --& --\\
Szeliski and Golland~\cite{szeliski1998stereo} & $\checkmark$ & --& --& --& --& --& --& --& --& --& --& --& --& --& --& --& --& --& --\\
Uras et al.~\cite{uras1988computational}& --& --& --& --& --& --& $\checkmark$ & --& --& --& --& --& --& --& --& --& --& --& --\\
GC-MVSNet~\cite{Vats2024GCMVSNet}  & $\checkmark$ & --& $\checkmark$ & --& --& --& --& --& --& --& --& --& --& --& --& --& --& --& --\\
Wang et al.~\cite{wang2020geometric} & --& --& --& --& --& $\checkmark$ & --& --& --& --& --& --& --& --& --& --& --& --& --\\
Wang et al.~\cite{wang2004image} & --& --& --& --& --& $\checkmark$ & --& --& --& --& --& --& --& --& --& --& --& --& --\\
Motif-GCNs~\cite{wen2022motif} & --& --& --& --& --& $\checkmark$ & --& --& --& --& --& --& --& --& --& --& --& --& --\\
Xu et al.~\cite{xu2022semi}  & --& $\checkmark$ & --& --& --& --& --& $\checkmark$ & --& --& --& --& --& --& $\checkmark$ & --& --& --& --\\
Xu et al.~\cite{xu2021self}  & --& $\checkmark$ & --& --& --& --& --& --& --& --& --& --& --& --& --& --& $\checkmark$ & $\checkmark$ & --\\
Xu et al.~\cite{xu2021digging}  & --& --& --& $\checkmark$ & --& --& --& --& --& --& --& --& --& --& --& --& --& --& --\\
Yang et al.~\cite{yang2021self} & --& --& --& --& --& --& --& --& --& --& --& --& --& --& --& $\checkmark$ & --& --& --\\
Yang et al.~\cite{yang2023geometry}& $\checkmark$ & --& --& --& --& --& $\checkmark$ & --& --& --& --& --& --& --& --& --& --& --& --\\
Yang et al.~\cite{yang2017unsupervised}  & $\checkmark$ & --& --& --& $\checkmark$ & --& $\checkmark$ & $\checkmark$ & --& --& --& $\checkmark$ & --& --& --& --& --& --& --\\
MVSNet ~\cite{yao2018mvsnet} & $\checkmark$ & --& --& --& --& --& --& --& --& --& --& --& --& --& --& --& --& --& --\\
Yin et al.~\cite{yin2019enforcing} & --& --& --& --& --& --& --& --& --& --& --& $\checkmark$ & --& --& --& --& --& --& --\\
GeoNet~\cite{Yin2018GeoNet} & --& --& --& --& --& $\checkmark$ & $\checkmark$ & --& --& --& --& --& --& --& --& --& --& --& --\\
p$^2$Net~\cite{yu2020p} & --& $\checkmark$ & --& --& --& --& $\checkmark$ & --& $\checkmark$ & --& --& --& --& --& --& --& --& --& --\\
Zhao et al.~\cite{zhao2022robust}  & --& $\checkmark$ & --& --& --& --& --& --& --& --& --& --& --& --& --& --& --& --& --\\
Zhao et al.~\cite{zhao2019geometry} & --& --& --& --& --& --& --& --& --& --& --& --& --& --& $\checkmark$ & --& --& --& --\\
Zhu et al.~\cite{zhu2021multi} & --& --& --& --& --& --& --& --& --& --& --& --& --& $\checkmark$ & --& --& --& --& --\\
Zhu et al.~\cite{zhu2021deep} & $\checkmark$ & --& --& --& --& --& --& --& --& --& --& --& --& --& --& --& --& --& --\\
\bottomrule
\end{tabular}
}}
\caption{Most methods use multiple constraints spanning various sub-categories, problem types, 
and datasets, meaningful qualitative or quantitative comparisons are difficult. 
This table highlights the use of sub-constraints across all methods providing a clear categorization.}
\label{table:methods-and-concept}
\end{center}
\end{table}

This survey is organized in $8$ sections. Starting from Sec. \ref{sec:plane-sweep}, 
we discuss the broad classification of geometric constraints presented in our taxonomy 
shown in
Fig. \ref{fig:taxonomy}. For most of the sections, we first describe 
the most common mathematical formulation of the geometry-inspired concept that covers the majority
of the methods, and then we describe different modifications applied 
to it by specific methods. 
Sec. \ref{sec:plane-sweep} describes the traditional plane sweep 
algorithm and its variants. 
Sec. \ref{sec:cross-view} focuses on all such geometric 
constraints that use alternate view(s) for enforcing consistency (cross-view consistency). 
Sec. \ref{sec:geometry-preserving} delves into geometric constraints that enforce 
structural consistency between a reference image and a target image to 
preserve the structural integrity of the scenes. 
Sec. \ref{sec:surface-normal} focuses on the orthogonal relation between 
depth and surface normal to guide geometric consistency. 
Sec. \ref{sec:geometric-attention} discusses the integration 
of geometric constraints in attention mechanism and Sec. \ref{sec:geometric-representation}
presents the methods to enforce geometry-based representation learning in 
deep neural networks. We present our conclusion in Sec. \ref{sec:conclusion}. 

\section{Plane Sweep Algorithm}
\label{sec:plane-sweep}
%%%%%%%%%%% list of methods %%%%%%%%%

Stereophotogrammetry is the process of estimating the 3D coordinates of 
points on an object by utilizing measurements from two or more images
of the object taken from different positions \cite{photogra_wiki}. 
This involves stereo matching where two or more images are used to 
find matching pixels in the images and to convert their 2D
positions into 3D depths \cite{szeliski2022computer}. 
The process of finding matching pixels is based on the 
\textit{geometry of stereo matching} (epipolar geometry), i.e the process of 
computing the range of possible locations of a pixel in one image 
that might appear in another image. In this section, we first discuss  
epipolar geometry for a pair of rectified images, and then describe a general resampling algorithm 
(\textit{plane sweep algorithm})  that can be used to perform multi-image stereo-matching
with arbitrary camera configurations. 

\begin{figure*}[t]
\begin{center}
    \includegraphics[width=1\textwidth]{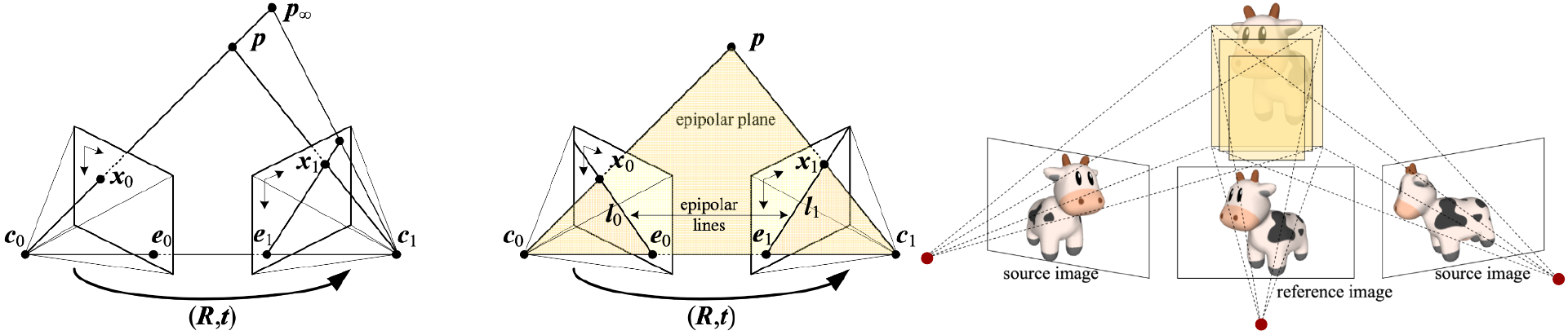}
    \caption{Epipolar geometry, Left: epipolar line segment corresponding to one ray (Fig. reprinted from \cite{szeliski2022computer}), 
    Center: the corresponding set of epipolar lines and their epipolar plane (Fig. reprinted from \cite{szeliski2022computer}), 
    $R$, $t$ are the rotation and translation parameters, and $l_i$ 
    is the epipolar line segment. Right: illustration of the plane-sweep algorithm (Fig. reprinted from\cite{zhu2021deep})}
    \label{fig:epipolar-plane-sweep}
\end{center}
\end{figure*}

Fig. \ref{fig:epipolar-plane-sweep} (left) shows the epipolar constraints 
of how a pixel in one image $x_0$ projects to an \textit{epipolar line segment} in the 
other image. The line segment is bounded by $p_{\infty}$ -- projection of the 
original viewing ray, and $c_0$ -- projection of the original camera center into
the second camera, called the \textit{epipole} $e_1$. The projections of the epipolar 
line in the second image back into the first image give us another line segment 
bounded by the corresponding epipole $e_0$. The extension of these two line 
segments to infinity gives us a pair of corresponding epipolar lines, 
Fig. \ref{fig:epipolar-plane-sweep}(center), that are the intersection 
of the two image planes, \textit{epipolar planes}, that pass through 
both camera centers $c_0$ and $c_1$. $p$ is the point of interest
\cite{hartley2003multiple, luong2001geometry}. 

Multi-image stereo reconstruction is a process to recover the 3D scene structure 
from multiple overlapping images with known intrinsic ($K$) and extrinsic ($E$)
camera parameters \cite{collins1996space}. More precisely, the photoconsistency is analyzed after 
projecting the source images to fronto-parallel planes of the reference camera frustum. 
The \textit{fronto-parallel} view of an image refers to the outcome 
of rectifying warped images. This allows the image to be viewed as it appears when 
directly in front of the observer's eyes (the reference camera is the observer). The reference camera 
frustum is a pyramid-shaped volume that determines what a reference camera can see 
and render in 3D space. The reference camera is at the top of the pyramid, and the frustum 
extends out in the direction the camera is looking, see Fig. \ref{fig:epipolar-plane-sweep} (right).
This process is commonly known as \textit{plane-sweep}
\cite{collins1996space, szeliski1998stereo, saito1999shape, zhu2021deep, Vats2024GCMVSNet}.

The plane-sweep method is based on the premise that areas of space 
where several image features' viewing rays intersect are likely to 
be the 3D location of observed scene features. In this method, 
a single plane partitioned into cells is swept through the volume 
of space along a line perpendicular to the plane \cite{collins1996space}, 
perpendicular to reference camera frustum
as shown in Fig. \ref{fig:epipolar-plane-sweep} (right). 
At each position of the plane along the sweeping path, the number of 
intersecting viewing rays is tallied. This is done by back-projecting
features from each source image onto the sweeping plane and noting 
the features that fall within some threshold of back-projected point 
position. Cells with the largest tallies are hypothesized as the location of 3D
scene features and the corresponding depth hypothesis (in the case of depth maps) are 
selected \cite{zhu2021deep, collins1996space}. 

The plane-sweep method can directly estimate the disparity (for stereo) 
or depth values (for MVS), but modern deep learning-based frameworks use it to create 
matching volume corresponding to each source image and the reference image. 
The matching volume is aggregated 
based on a metric to create a cost volume. 
The cost volume is then
regularized either by 3D-CNNs \cite{yao2018mvsnet, gu2020cascade, kendall2017end, ding2022transmvsnet} or 
by RNNs \cite{yao2019recurrent, peng2023rwkv}. 
The metric used for cost volume aggregation can vary from 
method to method. For example, Huang et al. \cite{huang2018deepmvs} compute pairwise 
matching costs between a reference image and neighboring source images and fuse them 
with max-pooling. MVSNet \cite{yao2018mvsnet}, R-MVSNet \cite{yao2019recurrent},  
and CasMVSNet \cite{gu2020cascade}
compute the variance 
between all encoded features warped onto each sweep plane and then regularize it with 
3D-UNet \cite{olaf2015unet}. TransMVSNet \cite{ding2022transmvsnet} computes similarity-based 
cost volume for regularization. Yang et al. \cite{yang2020cost} reuse cost 
volume from previous stage, along with the partial cost volume of the 
current stage in multi-stage MVS framework. They create a pyramidal 
structure for the cost volume.

Plane sweeping typically assumes surfaces to be fronto-parallel, which
causes ambiguity in matching where slanted surfaces are involved,
like urban scenes \cite{gallup2007real}. Gallup et al. \cite{gallup2007real}
propose to perform multiple plane sweeps, where each plane sweep 
is intended to take care of planar surfaces with a particular normal.
Their method is applied in three steps. First, the surface 
normals of the ground and facade planes are identified by 
analyzing 3D points obtained through the sparse structure 
from motion. Then, a plane sweep for each surface normal is 
applied, resulting in multiple depth hypotheses for each pixel 
in the final depth map. Finally, the best depth/normal combination 
for each pixel is selected based on a cost or by multi-label 
graph cut method \cite{boykov2001fast, greig1989exact, bishop2006pattern}.

\section{Cross-View Constraints}
\label{sec:cross-view}
%%%%%%%%%%% list of methods %%%%%%%%%

Cross-view constraints are applied to scenes with more than one view.
They can be applied to stereo (two views per scene)
and MVS ($N>2$ view per scene) frameworks by projecting one view, either 
reference or source view, to an other view or vice-versa. Once projected to 
the other view, various constraints, like photometric consistency, 
depth flow consistency, and view synthesis consistency, can be utilized. 
An alternate way of utilizing cross-view constraints is to use forward-backward 
reprojection, where one view is projected to the other view and then it is back-projected 
to the first view to check the geometrical consistency of the scene. 
All the sub-constraints in this section can be effectively used as part of an objective 
function during optimization (see Table \ref{table:organis-and-fit})
and has been observed to be effective in cases like cluttered background, 
repetitive patterns and texture-less regions \cite{mallick2020learning, zhao2022robust, xu2021digging, huang2021m3vsnet, liu2023geometric, khot2019learning, chen2021fixing, chen2019self, Vats2024GCMVSNet, xu2021digging}. 
In this section, 
we discuss all such approaches that use cross-view consistency constraints in 
end-to-end deep learning-based frameworks.

%%%%%%%%%%%%%%%%%%%%%%%%%%%%%%%%%%%%%%%%%%%%%%%%%%%%%%%%%%%%%%%%%%%%%%%%%%%%%%%%%%%%%%%%%%%%%%%%%%%%%%%%%%%%%%%%%%%
\subsection{Photometric Consistency}\label{subsec:photometric-consistency}

Photometric consistency minimizes the difference between a real image 
and a synthesized image from other views. The real and the synthesized
images are denoted differently for different tasks. For example, MVS tasks use the 
\textit{reference} ($I_{ref}$) and \textit{source} ($I_{src}$) 
images in place of the real and synthesized images, stereo tasks use 
\textit{left} ($I_{L}$) and \textit{right} ($I_{R}$) or vice-versa in place of the real and the synthesized images, and 
video depth estimation tasks use the next frame ($I_{+}$) and the current frame ($I_{0}$) in place of the real 
and the synthesized images.
For all these tasks, one view is warped to the other view using intrinsic ($K$) and 
extrinsic ($E$) camera parameters. The warping process brings both 
images in the same camera frustum for photometric consistency estimation.

Two main variations of photometric loss are \textit{pixel photometric loss} 
and \textit{gradient photometric loss}. As the name suggests, pixel
photometric loss is the comparison between pixel values of these images, and 
gradient photometric loss is the comparison of the gradients of these images.
Sometimes, pixel photometric loss and gradient photometric loss are combined 
for a more robust form of photometric loss, called \textit{robust photometric loss}. 

\begin{align}
    \label{eq:pixel-photometric}
    \mathcal{L}_{photo_{pixel}} &= \frac{||(I_{ref} - \hat{I}_{src \rightarrow ref}) \odot M||_{l}}{||M||_{1}}\\
    \label{eq:grad-photometric}
    \mathcal{L}_{photo_{grad}} &= \frac{||(\triangledown I_{ref} - \triangledown \hat{I}_{src \rightarrow ref}) \odot M||_{l}}{||M||_{1}}\\
    \label{eq:robust-photometric}
    \mathcal{L}_{photo_{robust}} &= \lambda_1 . \mathcal{L}_{photo_{pixel}} + \lambda_2 . \mathcal{L}_{photo_{grad}}
\end{align}

\noindent where $l$ denotes $L_1$ or $L_2$ norm, $M$ denotes the mask, and $\lambda_1, \lambda_2$ denote 
scaling factors for pixel and gradient photometric losses, respectively. $\odot$ denotes pixel-wise 
multiplication. 

There are different ways of formulating Eqs \ref{eq:pixel-photometric}, \ref{eq:grad-photometric} 
and \ref{eq:robust-photometric} based on the choice of view to be warped. Most methods 
\cite{mallick2020learning, zhao2022robust, xu2021digging, huang2021m3vsnet, liu2023geometric, khot2019learning} 
warp source views to the reference view, or \textit{source-reference} warp, 
as shown in Eqs \ref{eq:pixel-photometric}, \ref{eq:grad-photometric} 
and \ref{eq:robust-photometric}. But some methods 
warp the reference view to source views (\textit{reference-source} warp)\cite{li2022ds}, 
by replacing \textit{src} with \textit{ref} and vice-versa 
in Eqs \ref{eq:pixel-photometric}, \ref{eq:grad-photometric} and \ref{eq:robust-photometric}.
Other methods like Dong et al. \cite{dong2022patchmvsnet} use patch-wise photometric 
consistency to estimate the photometric loss. 
We highlight such variations of 
photometric consistency formulation in the next few paragraphs.

%% pixel photometric loss
We start with pixel photometric loss and its variations.
Mallick et al. \cite{mallick2020learning} use pixel formulation of photometric loss, Eq. \ref{eq:pixel-photometric}, 
between reference and source views to enforce geometric consistency in a self-supervised MVS method. 
Zhao et al. \cite{zhao2022robust} use a similar formulation of pixel photometric loss 
in self-supervised monocular depth estimation problem to promote cross-view 
geometric consistency.
Xu et al. \cite{xu2022semi} use a slightly different formulation of pixel photometric loss based 
on the condition that a pixel in one view finds a valid pixel in another view for semi-supervised 
MVS method.

\begin{equation}
    \mathcal{L}_{photo_{pixel}} = \frac{\Phi(1 \leq \hat{p}_{i} \leq [H,W]) || I_{ref}(p_i) - \hat{I}_{src \rightarrow ref}(p_i) ||_{L_2}}{\Phi(1 \leq \hat{p}_{i} \leq [H,W])},
    \label{eq:pixel-photoe-occlusion}
\end{equation}

\noindent where $p_i$ denotes a pixel, $\Phi(1 \leq \hat{p}_{i} \leq [H,W])$ indicates 
whether the pixel $p_i$ can find a valid pixel $\hat{p}_i$ in 
other source view. $H, W$ denote the height and width of the image, respectively.

 Li et al. \cite{li2022ds} use a slightly different formulation of pixel photometric loss by warping 
reference view to source views, instead of warping source view to the reference view, to calculate the
$L_1$ distance between reference-source depth maps, 

\begin{equation}
    \mathcal{L}_{photo_{pixel}} = \frac{1}{N-1} \sum^{N}_{i=1}|I_{src_i} - \hat{I}_{ref \rightarrow src_i}|
\end{equation}

\noindent where $N$ is the total number of source views.

All the above-mentioned methods use pixel-wise warping operations to estimate photometric error. 
Yu et al. \cite{yu2020p} propose patch-based warping of the extracted key points. They use the 
point selection strategy from Direct Sparse Odometry (DSO) \cite{engel2018DSO} and define a 
support domain $\Omega(P_i)$ over each point $P_i$'s local window. The photometric 
error is then estimated over each support domain $\Omega(P_i)$, instead of a single point.
This is called \textit{patch photometric consistency}.
Dong and Yao \cite{dong2022patchmvsnet} applied a similar approach to estimate 
patch photometric error. Unlike \cite{yu2020p}, which uses DSO to extract key points, 
they use each pixel as a key point. They define a 
square patch centered on the pixel $P$ as $\Omega(P)$. The local patch 
$\Omega(P)$ is small so that it can be treated as if it lies in a plane \cite{yu2020p} and  
assumed to share the same depth as the center pixel. This patch 
is warped from the source view to the reference view and the $L_1$ distance between pixel 
values is estimated as $\mathcal{L}_{photo_{patch}}$

\begin{equation}
    \mathcal{L}_{photo_{patch}} = \frac{1}{N-1} \sum^{N}_{i=1}|I_{ref} - \hat{I}_{src \rightarrow ref}| \odot M_{ref},
    \label{eq:photo-patch}
\end{equation}

\noindent where $N$ indicates the number of source views and $M_{ref}$ indicates the reference view mask. $\odot$ denotes 
element-wise multiplication.

%% Robust photometric loss

While pixel photometric consistency is widely used to achieve geometric 
consistency across views, their performance is susceptible to changes in lighting conditions. 
Change in lighting conditions makes enforcing pixel-level photometric consistency difficult, but 
image gradients are more invariant to such changes. Many methods employ gradient 
photometric loss alongside pixel photometric loss. Since the addition of 
the gradient term makes the photometric loss more robust, it is called \textit{robust photometric loss}. 
MVS methods like 
\cite{xu2021digging, huang2021m3vsnet, liu2023geometric, khot2019learning, xu2021self} 
use this formulation, Eq. \ref{eq:robust-photometric}, during end-to-end training. 
$\lambda_1$ and $\lambda_2$ are the tunable hyper-parameters in Eq. \ref{eq:robust-photometric}.
Robust photometric loss also helps with regions of low-texture, repetitive patterns and cluttered background by 
double checking the pixel consistency across multiple views.

While most mentioned MVS methods use an asymmetric pipeline, estimating only the reference depth map
using both the reference and source RGB images, 
Dai et al. \cite{dai2019mvs2} use a symmetric pipeline for MVS, i.e. the network 
predicts depth maps of all views simultaneously. With $N$ depth estimates, one per view,
the method uses a bidirectional calculation of photometric consistency 
between each pair of views, called \textit{cross-view consistency loss}. They do not 
use robust formulations for cross-view consistency.

%%%%%%%%%%%%%%%%%%%%%%%%%%%%%%%%%%%%%%%%%%%%%%%%%%%%%%%%%%%%%%%%%%%%%%%%%%%%%%%%%%%%%%%%%%%%%%%%%%%%%%%%%%%%%%%%%%%
\subsection{Geometric Consistency}\label{subsec:geometric-consistency}

Like photometric consistency, geometric consistency also involves 
cross-view consistency checks with projections. For photometric consistency, 
one view, either reference or source, is warped to another view to calculate 
the consistency error. Geometric consistency employs \textit{forward-backward reprojection} (FBR)
to estimate the error. FBR involves a series of projections of the reference view
to estimate the geometric error, as shown in Alg. \ref{alg:forward-backward-reprojection}. 
First, we project the reference image ($I_{ref}$) to the source view ($I_{ref \rightarrow src}$), then 
we remap the projected reference view $I_{ref \rightarrow src}$ to generate $I_{{src}_{remap}}$, and finally 
we back-project $I_{{src}_{remap}}$ to the reference view to obtain $I_{ref \rightleftarrows src}$. 
$I_{ref \rightleftarrows src}$ is then compared with the original $I_{ref}$ to estimate the photometric error.

\begin{algorithm}[t]
\footnotesize
\begin{algorithmic}
    \State \textbf{Inputs:} $I_{ref}: \text{ref. image},c_{ref}: \text{ref. camera params.},I_{src}: \text{src. image},c_{src}: \text{src. camera params.}$
    \State $K_{ref}, E_{ref} \gets c_{ref}$;  $K_{src}, E_{src} \gets c_{src}$
    \State $I_{(ref \rightarrow src)} \gets K_{src} \cdot E_{src} \cdot E_{ref}^{-1} \cdot K_{ref}^{-1} \cdot I_{ref}$ \Comment{Project}
    \State $X_{I_{(ref \rightarrow src)}}, Y_{I_{(ref \rightarrow src)}} \gets I_{(ref \rightarrow src)}$ 
    \State $I_{{src}_{remap}}\gets REMAP(I_{src}, X_{I_{(ref \rightarrow src)}}, Y_{I_{(ref \rightarrow src)}})$  \Comment{Remap}
    \State $I_{ref \rightleftarrows src} \gets K_{ref} \cdot E_{ref} \cdot E_{src}^{-1} \cdot K_{src}^{-1} \cdot I_{{src}_{remap}}$         \Comment{Back project}
\end{algorithmic}
\caption{Forward Backward Reprojection (FBR)}
\label{alg:forward-backward-reprojection}
\end{algorithm}

Dong and Yao \cite{dong2022patchmvsnet} use cross-view geometric consistency 
by applying FBR in the pixel domain in an unsupervised MVS pipeline. Once the FBR steps are done, 
the actual pixel values between the original reference image $I_r$ and back-projected reference images 
$I_{r \rightleftarrows s}$ are used to check the geometrical consistency of the depth estimates, which diminishes the 
matching ambiguity between reference and source views,

\begin{equation}
    \mathcal{L}_{geometric} = \frac{1}{N} \sum^{N}_{i=1} |I_{ref} - \hat{I}^{i}_{ref \rightleftarrows src}| \odot M_{ref} ,
\end{equation}

\noindent where $N$ indicates the number of source views and $M_{ref}$ denotes the reference view mask. $\odot$ is the 
element-wise multiplication operation.

Geometric consistency can be extended to the 3D coordinates of the camera using the  
concept of back-projection \cite{chen2021fixing, chen2019self, Vats2024GCMVSNet}. FBR has 
also been found helpful in images with low-texture regions, occlusion and repetitive patterns \cite{Vats2024GCMVSNet, chen2019self}.
Chen et al. \cite{chen2019self} apply this modified geometric
consistency in 3D space for a video depth estimation problem, called \textit{3D geometric consistency}. 
For a given source image pixel $P_{src}$ and corresponding reference image pixel $P_{ref}$, their 
3D coordinates are obtained by back projection. The inconsistency between the estimates of the same 
3D point is then estimated and used as a penalty. The loss value represents the 3D discrepancy 
of predictions from two views. Chen et al. \cite{chen2021fixing} use a similar formulation to overcome 
the deficiencies of photometric loss in a self-supervised monocular depth estimation framework.

\begin{figure*}[t]
\begin{center}
    \includegraphics[width=0.5\textwidth]{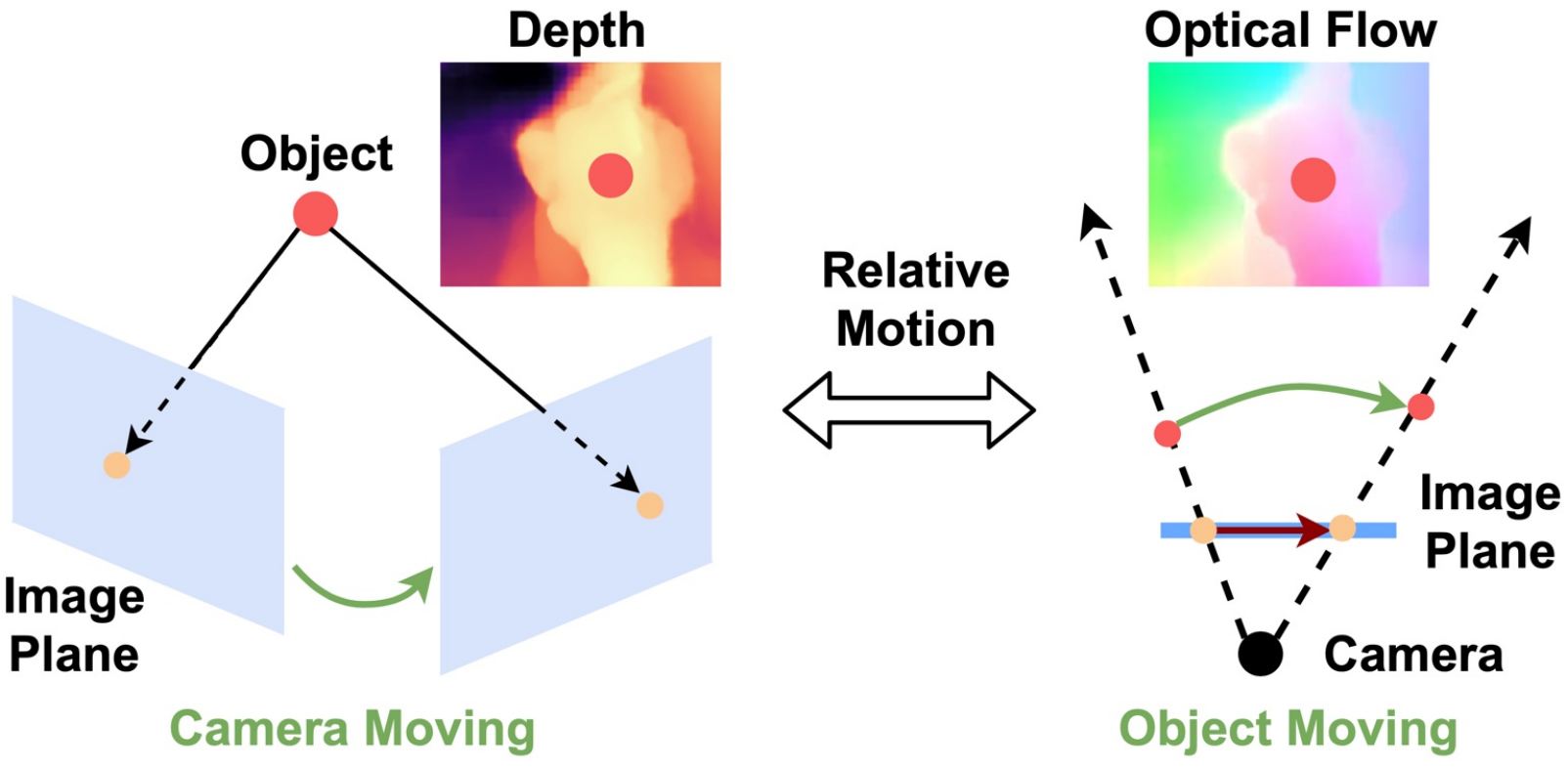}
    \caption{Intuition of Depth2Flow module. The relative motion of 
    a moving camera can be viewed as a moving object and a still camera to 
    estimate optical flow. Figure reprinted from \cite{xu2021digging}}
    \label{fig:depth-flow-motion}
\end{center}
\end{figure*}
%%%%%%%%%%%%%%%%%%%%%%%%%%%%%%%%%%%%%%%%%%%%%%%%%%%%%%%%%%%%%%%%%%%%%%%%%%%%%%%%%%%%%%%%%%%%%%%%%%%%%%%%%%%%%%%%%%%
\subsection{Cross-View Depth-Flow Consistency}\label{subsec:flow-consistency}

Depth-flow estimations are typically used in optical flow problems \cite{liu2020learning, meister2018unflow}. 
But they can easily be adapted to MVS problems by estimating flow from estimated depth maps 
as well as from input RGB images and comparing them. Xu et al. \cite{xu2021digging} propose a 
novel flow-depth consistency loss to regularize the ambiguous supervision in the foreground of depth maps. 
Estimation of flow-depth consistency loss requires two modules for an MVS method, RGB2Flow and Depth2Flow. 
As the name suggests, the Depth2Flow module transforms the estimated depth maps to virtual optical flow fields between 
the reference and arbitrary source view and the RGB2Flow module uses \cite{liu2020learning} to predict the
optical flow from the corresponding reference-source pairs. The two predicted flows should be consistent 
with each other. 

\paragraph{Depth2Flow module:}
In an MVS system, cameras move around the object to collect multi-view images. If we consider 
the relative motion between the object and the camera, the camera can be assumed to be fixed
and the object can be assumed to be in motion toward the virtually still camera, see Fig. \ref{fig:depth-flow-motion}. 
This virtual motion can be represented as a dense 3D optical flow and should be 
consistent with the 3D correspondence in real MVS systems. The virtual flow for a pixel $p_i$
can be defined as,

\begin{equation}
    \hat{F}_{ref \rightarrow src} = Norm[K_{src}.E_{src}(K_{ref}.T_{ref})^{-1} I_{ref}(p_i)] - p_i ,
\end{equation}

\paragraph{RGB2Flow module:}
To estimate the optical flow from RGB input images, a pre-trained model can be used. Xu et al. \cite{xu2021digging}
use \cite{liu2020learning} to estimate the forward flow (reference to source image flow, $F_{ref \rightarrow src}$) 
and backward-flow (source to reference view flow, $F_{src \rightarrow ref}$). 
Each source view is combined with one reference view to form a pair to obtain a forward and backward flow i.e.
$F_{ref \rightarrow src}$ and $F_{src \rightarrow ref}$. The consistency of both the 
flows ($F_{ref \rightarrow src}$ and $F_{src \rightarrow ref}$)
are compared with the virtual flow ($\hat{F}_{ref \rightarrow src}$) to estimate
the flow-depth consistency loss. The occluded pixels from different source views are masked out with 
mask $M_{ref \rightarrow src}$ and the error is given as,

\begin{align}
    \label{eq:flow-mask}
    M_{ref \rightarrow src} &= \{|F_{ref \rightarrow src} + F_{src \rightarrow ref}| > \epsilon \} \\
    \label{eq:flow-error}
    \mathcal{L}_{flow} &= \sum^{HW}_{i=1} \genfrac{}{}{0pt}{}{min}{2 \leq s_j \leq V}  \frac{|| F_{ref \rightarrow {src}_j}(p_i) - \hat{F}_{ref \rightarrow {src}_j}(p_i).M_{}(p_i) ||_2}{\sum^{HW}_{i=1} M_{ref \rightarrow {src}_j}(p_i)} ,
\end{align}

\noindent where $\epsilon$ is a threshold set to 0.5, and $H$ and $W$ are the height and width of the image. Instead of averaging 
the difference between $F_{ref \rightarrow src}$ and $\hat{F}_{ref \rightarrow src}$ on all source views, a minimum error 
is estimated at each pixel, see Eq \ref{eq:flow-error}. Godard et al. \cite{godard2019digging} 
introduced the minimum error calculation 
to reject occluded pixels in depth estimation. Both the modules, 
Depth2Flow and RGB2Flow, are fully differentiable 
and can be used in end-to-end training setups. 

Chen et al. \cite{chen2019point} use point-flow information to refine the 
estimated depth map in an MVS framework. Using extracted features from the 
input images and the previous stage's estimated depth map, they generate 
a feature augmented point cloud and use a \textit{point-flow} module 
to learn the depth residual to refine the previously estimated depth map.
The point-flow module estimates the displacement of a point to the ground 
truth surface along the reference camera direction by observing the 
neighboring points from all views. 

%%%%%%%%%%%%%%%%%%%%%%%%%%%%%%%%%%%%%%%%%%%%%%%%%%%%%%%%%%%%%%%%%%%%%%%%%%%%%%%%%%%%%%%%%%%%%%%%%%%%%%%%%%%%%%%%%%%
\subsection{View Synthesis Consistency}\label{subsec:view-synthesis}

\begin{figure*}[t]
\begin{center}
    \includegraphics[width=0.5\textwidth]{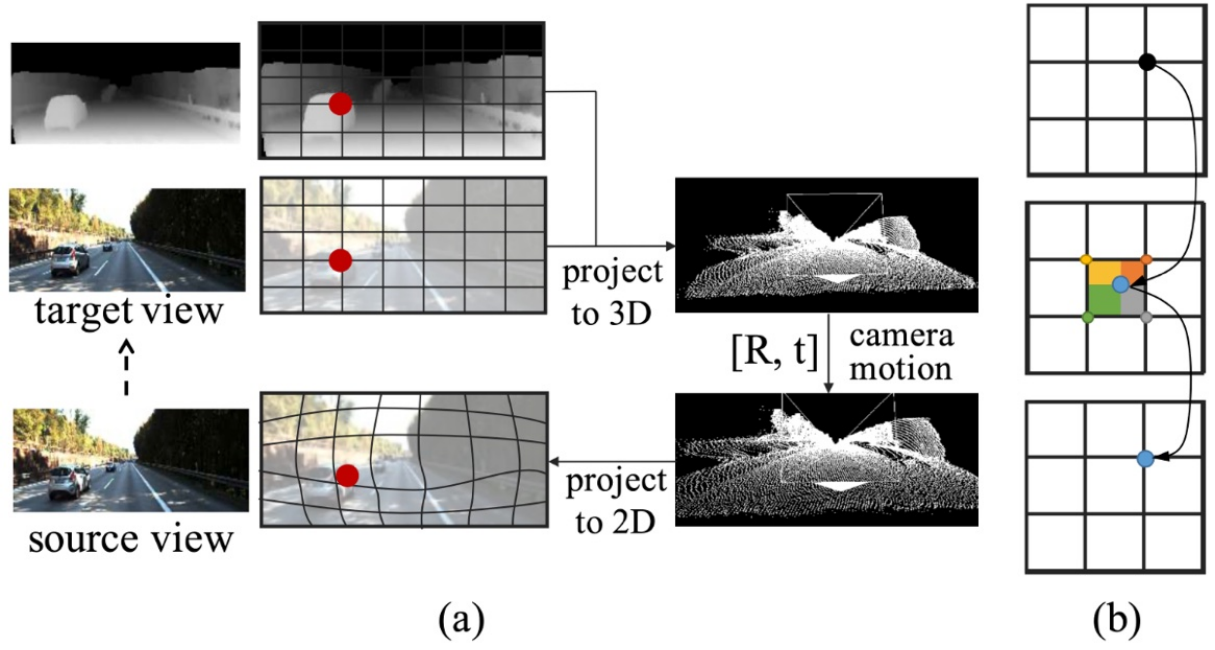}
    \caption{(a) Steps of view synthesis using RGB image and its depth map.
    (b) shows a bilinear interpolation step for adjusting values to pixel space. R and t are rotation
    and translation parameters.
    Figure reprinted from \cite{yang2017unsupervised}}
    \label{fig:view-synthesis}
\end{center}
\end{figure*}

Most of the monocular, stereo, and MVS depth estimation frameworks use ground truth as a 
supervision signal. While these frameworks may utilize the additional source view 
images in the pipeline, they always estimate only one depth map, the reference view depth map.
Estimation and use of only one depth map may not provide enough geometric information 
for consistent estimation. To address this gap, many methods 
\cite{bauer2021nvs, yang2017unsupervised, li2022ds, dai2019mvs2}
synthesize an additional view (see Fig. \ref{fig:view-synthesis}), either a
depth map or an RGB image (commonly referred to as target view), using camera parameters and 
reference view information. This additional view, when included in the training framework, 
provides additional geometric consistency information during the learning process. 

Bauer et al. \cite{bauer2021nvs} use view synthesis in a monocular depth estimation framework. 
They apply two networks, depth network (DepNet) and synthesis network (SynNet) in a series of 
operations to enforce geometric constraints with view synthesis. 
First, the RGB input (source view) is used in DepNet to generate a corresponding 
depth estimate. The estimated depth map is projected to a target view and, using
SynNet, the holes in the target view are filled. Finally, the synthesized RGB target 
view is used as input to DepNet to estimate its depth map. This ensures that 
the DepNet learns to estimate geometrically consistent depth estimates of both 
the source and the synthesized target view. They use an $L_1$ loss to enforce consistency.

Yang et al. \cite{yang2017unsupervised} also synthesize RGB target view to improve 
geometric consistency in video depth estimation. With estimated pixel matching pairs 
between source ($I_s$) and target views ($I_t$), they synthesize a target view $\hat{I}_s$
using the source view, camera parameters, and bilinear interpolation \cite{garg2016unsupervised}.
To handle occlusion and movement of objects, an explainability mask ($M_s$) is applied 
during the loss calculation,

\begin{equation}
    \mathcal{L}_{ViewSynthesis} = \sum^{S}_{s=1} |I_t - \hat{I}_s| \odot M_s ,
    \label{eq:view-synthesis-yang}
\end{equation}

\noindent where $s$ are the source views. 
 
Typically, MVS methods regularize a cost volume that is generated using multiple 
source views, to estimate the reference view depth map. 
Li et al. \cite{li2022ds} argue that by estimating only the reference 
view depth map from the cost volume, an MVS method underutilizes the information present in the 
cost volume. They propose a module to synthesize source depths
from the cost volume by projecting the pixels in the reference view at different 
depth hypotheses to the source views. They use robust photometric consistency 
(Sec. \ref{subsec:photometric-consistency}) to estimate the view synthesis loss.

Dai et al. \cite{dai2019mvs2} synthesize source view depth maps using the reference 
view depth maps in the MVS pipeline for additional geometric consistency. For each 
pair of reference-source views, they calculate the bidirectional error, i.e. 
$\mathcal{L}_{ref \rightarrow src}$ and $\mathcal{L}_{src \rightarrow ref}$ along 
with $\mathcal{L}_{smooth_{\triangledown}}$ and $\mathcal{L}_{smooth_{\triangledown^2}}$ (Sec. \ref{subsec:smooth}) to estimate 
$\mathcal{L}_{ViewSynthesis}$. They use a robust formulation of photometric consistency loss 
(Sec. \ref{subsec:photometric-consistency}) to estimate $\mathcal{L}_{ref \rightarrow src}$ and $\mathcal{L}_{src \rightarrow ref}$ 
and add structural similarity (SSIM) error term
(Sec. \ref{subsec:SSIM}) in the loss, 

\begin{align}
    \label{eq:bidirectional-view-synthesis}
    \mathcal{L}_{ViewSynthesis}^{ref \rightarrow src} &= (\mathcal{L}_{ref \rightarrow src} + \mathcal{L}_{src \rightarrow ref}) + (\mathcal{L}^{ref}_{smooth_{\triangledown}} + \mathcal{L}^{ref}_{smooth_{\triangledown^2}})\\
    \label{eq:bidirectional}
    \mathcal{L}_{ref \rightarrow src} &= \mathcal{L}_{photo_{robust}} + \mathcal{L}^{ref}_{SSIM}\\
    \mathcal{L}_{src \rightarrow ref} &= \mathcal{L}_{photo_{robust}} + \mathcal{L}^{src}_{SSIM}
\end{align}

\noindent where $ref, src$ are the reference and the source views. We describe the other loss terms in Sec. \ref{sec:geometry-preserving}.

%%%%%%%%%%%%%%%%%%%%%%%%%%%%%%%%%%%%%%%%%%%%%%%%%%%%%%%%%%%%%%%%%%%%%%%%%%%%%%%%%%%%%%%%%%%%%%%%%%%%%%%%%%%%%%%%%%%

\section{Geometry Preserving Constraints}
\label{sec:geometry-preserving}
%%%%%%%%%%% initial content %%%%%%%%%

Besides cross-view consistency constraints, there are other 
ways of enforcing structural consistency in a depth estimation pipeline. 
In this section, we discuss all such approaches that utilize alternative 
methods of enforcing geometric constraints. We have classified these  
methods into four broader categories, i.e. structural similarity index measurement (SSIM), 
edge-aware smoothness constraints, consistency regularization, and 
planar consistency. All these constraints can help with problems like 
cluttered background, repetitive patterns or texture-less regions. 
They can also be integrated within an objective function for enhanced geometric learning.
We discuss each of these approaches in detail and highlight 
research that uses these methods in their pipeline.

%%%%%%%%%%%%%%%%%%%%%%%%%%%%%%%%%%%%%%%%%%%%%%%%%%%%%%%%%%%%%%%%%%%%%%%%%%%%%%%%%%%%%%%%%%%%%%%%%%%%%%%%%%%%%%%%%%%
\subsection{Structural Similarity Index Measurement}\label{subsec:SSIM}

\begin{figure*}[t]
\begin{center}
    \includegraphics[width=0.65\textwidth]{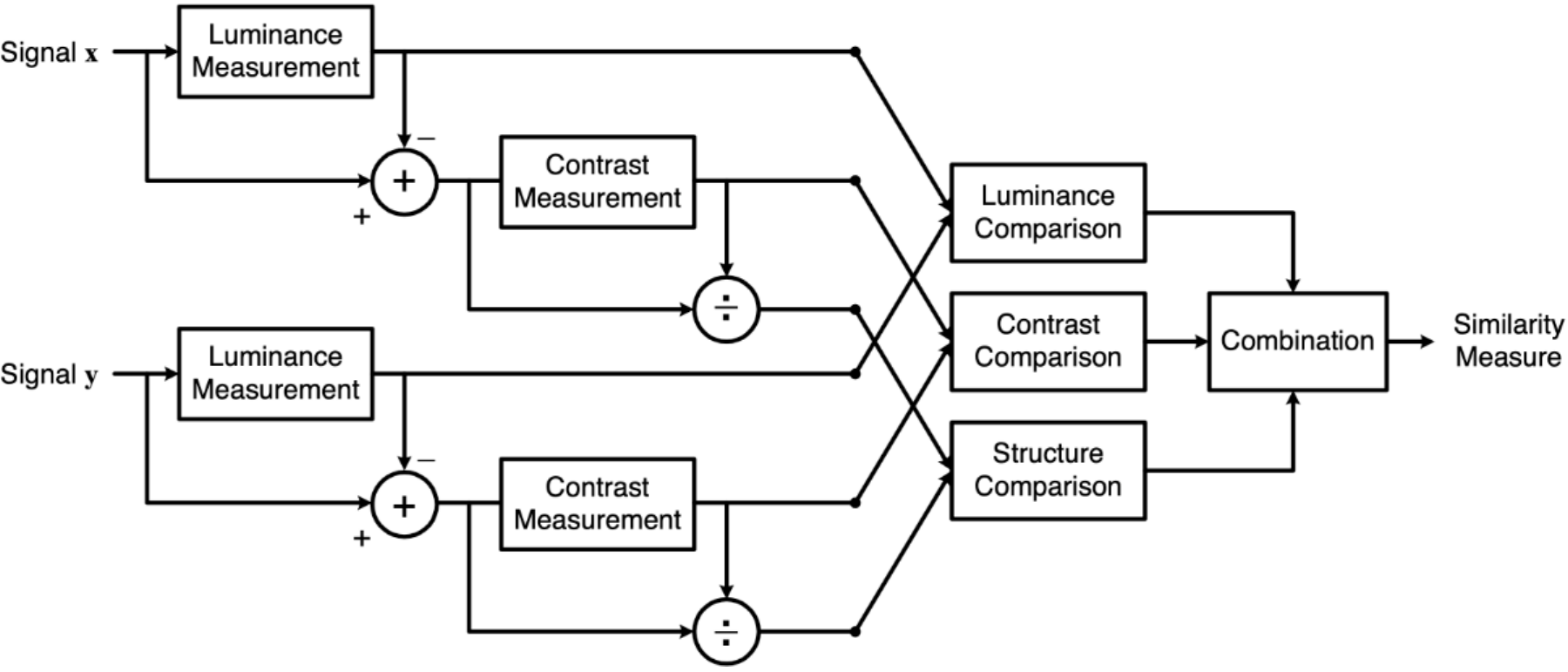}
    \caption{SSIM measurement system as described in \cite{wang2004image}. Figure reprinted from \cite{wang2004image}}
    \label{fig:ssim}
\end{center}
\end{figure*}

Structural similarity index measurement (SSIM) 
is a full-reference image quality assessment technique \cite{wang2004image}. 
Its assessment is based on the degradation of structural information between the
reference and noisy image. Specifically, it compares local patterns of 
pixel intensities that have been normalized for luminance and contrast. 
Luminance of a surface is the product of its illumination and reflectance, but 
the structure of the object is independent of illumination. SSIM separates
the influence of illumination to analyze the structural information 
in an image \cite{wang2004image,eskicioglu1995image, zhao2015loss, mallick2020learning, wen2022motif, dong2022geometry}. 

Objective image quality metrics can be roughly classified into three categories
based on the availability of distortion-free (original or reference) images. 
The metric is called \textit{full-reference} when the complete reference 
image is known, \textit{no-reference} when the reference image 
is not available, and \textit{reduced-reference} when the reference 
image is partially available. Eskicioglu and Fisher \cite{eskicioglu1995image}
discuss several such image quality metrics and their performance. 

Wang et al. \cite{wang2004image} define structural information as the 
attributes that represent the structural information of objects in an image, 
independent of average luminance and contrast. As shown in Fig. \ref{fig:ssim}, 
given two aligned images, X and Y, one 
of which is assumed to be the reference quality, we can have a quantitative 
measurement of the second signal with SSIM. Three separate tasks are considered 
for structural similarity measurement: luminance, contrast, and structure \cite{wang2004image}.

The mathematical formulation of SSIM, 

\begin{align}
    \label{eq:ssim}
    SSIM(X,Y) &= [l(X,Y)]^\alpha  [c(X,Y)]^\beta  [s(X,Y)]^\gamma; \alpha > 0, \beta > 0, \gamma > 0 \\
    \label{eq:luminance}
    l(X,Y) &= \frac{2\mu_x \mu_y + c_1}{\mu_x^2 + \mu_y^2 + c_1} ; c_1 = (K_1L)^2, K_1 \ll 1\\
    \label{eq:contrast}
    c(X,Y) &= \frac{2\sigma_x \sigma_y + c_2}{\sigma^2_x + \sigma^2_y + c_2}; c_2 = (K_2L)^2, K_2 \ll 1\\
    \label{eq:structure}
    s(X,Y) &= \frac{\sigma_{xy} + c_3}{\sigma_x \sigma_y + c_3}, c_3 = \frac{c_2}{2} ,
\end{align}

\noindent is a scaled product of the three components. Where the luminance  function $l(X, Y)$ is a function of mean intensities 
$\mu_x$ and $\mu_y$ as shown in Eq. \ref{eq:luminance}. The contrast comparison $c(X,Y)$
is a function of $\sigma_x$ and $\sigma_y$ as shown in Eq. \ref{eq:contrast} and the 
structure  measure $s(X,Y)$ is a function of correlation coefficient $\sigma_{xy}$, 
$\sigma_x$ and $\sigma_y$ as shown in Eq. \ref{eq:structure}. $L$ is the dynamic range 
of the pixel values (255 for grayscale image). The mean, standard deviation, and correlation coefficient 
of the signals are calculated using Eqs. \ref{eq:mean-signal}, \ref{eq:std-signal}, and \ref{eq:coor-coef}. 
Where $N$ is the total number of pixels in an image.

\begin{align}
    \label{eq:mean-signal}
    & \mu_x = \frac{1}{N} \sum^{N}_{i=1} x_i ; \mu_y = \frac{1}{N} \sum^{N}_{i=1} y_i\\
    \label{eq:std-signal}
    & \sigma_x = (\frac{1}{N-1} \sum^{N}_{i=1} (x_i - \mu_x)^2)^{\frac{1}{2}} ; \sigma_y = (\frac{1}{N-1} \sum^{N}_{i=1} (y_i - \mu_y)^2)^{\frac{1}{2}}\\
    \label{eq:coor-coef}
    & \sigma_{xy} = \frac{1}{N-1} \sum^{N}_{i=1} (x_i - \mu_x)(y_i - \mu_y) 
\end{align}

SSIM should be computed locally rather than globally for several reasons. First, 
image statistical features are highly spatially non-stationary i.e. statistical features
are not the same across different spatial locations within an image. Second, image distortions 
may be space-variant, and third, local quality measurement delivers more information about 
the quality degradation by providing a spatially varying quality map of the image.

Wang et al. \cite{wang2004image} defined $SSIM$ for 
two signals in the same domain, i.e.
it can estimate the structural similarity between two RGB or grayscale images or
between two depth maps or two patches. In an end-to-end framework, 
where we minimized the loss, we 
want to maximize $SSIM$ for better result. 
Since the upper bound of $SSIM$ is $1$,
we can instead minimize $ 1 - SSIM$,

\begin{equation}
    \mathcal{L}_{SSIM} = \frac{1}{N} \sum^{N}_{i=1} \frac{1 - SSIM(X, Y)}{K} \odot M
    \label{eq:ssim-general-mvs}
\end{equation}

\noindent where $N$ is the number of source views, and $X$ and $Y$ 
are the two images to be compared. $M$ is the mask to handle occlusion 
and $K$ is a constant. 

Zhao et al. \cite{zhao2015loss} use this formulation
with no mask and $K=1$ for image restoration problems. They calculate the means and 
standard deviations (Eqs. \ref{eq:mean-signal}, \ref{eq:std-signal} and \ref{eq:coor-coef}) 
using a Gaussian filter with standard deviation $\sigma_{G}$. The choice of $\sigma_{G}$
impacts the quality of the processed results. With smaller values of $\sigma_G$ the 
network loses the ability to preserve local structures which introduces
artifacts in the image, and for larger $\sigma_G$, the network preserves noises around the edges. Instead of 
finetuning the value of $\sigma_G$, Zhao et al. \cite{zhao2015loss} propose a 
multi-scale formulation of SSIM (MS-SSIM), where all results from the variations of 
$\sigma_G$ are multiplied together. 

Inherently, both MS-SSIM and SSIM are not particularly sensitive to the change of 
brightness or shift of colors. However, they preserve the contrast in high-frequency
regions. $L_1$ loss, on the other hand, preserves colors and luminance - an error 
weighted equally regardless of the local structure - but does not produce the same impact 
for contrast. For best impact, Zhao et al. \cite{zhao2015loss} combines both these
terms, 

\begin{equation}
    \mathcal{L}^{mix}_{SSIM} = \alpha . \mathcal{L}_{SSIM} + (1 - \alpha) . \mathcal{L}^{l} ,
    \label{eq:mix-ssim}
\end{equation}

\noindent where, $\alpha$ is a tunable hyper-parameter and $l$ is $L_1$ or $L_2$ norm.

Mallick et al. \cite{mallick2020learning} use the simplest form of $\mathcal{L}_{SSIM}$ (Eq. \ref{eq:ssim-general-mvs})
in a self-supervised MVS framework with no mask $M$, $X=I_{ref}$, $Y=\hat{I}_{src}$ and $K=1$.
Huang et al. \cite{huang2021m3vsnet} use $\mathcal{L}_{SSIM}$ (Eq. \ref{eq:ssim-general-mvs})
with $X=I_{ref}$, $Y=\hat{I}_{src}$ and $K=2$ in unsupervised MVS framework. Mahjourian et al. 
\cite{mahjourian2018unsupervised} also use it in unsupervised MVS framework with $K=1$ (Eq. \ref{eq:ssim-general-mvs}), 
$c_1={0.01}^2$ and $c_2={0.03}^2$ (Eqs. \ref{eq:luminance} and \ref{eq:contrast}). 
Khot et al. \cite{khot2019learning} use the same formulation as \cite{mahjourian2018unsupervised} in an 
unsupervised MVS framework. Li et al. \cite{li2022ds} use bidirectional calculation, 
forward ($\mathcal{L}^{forward}_{SSIM}$) with $X=I_{ref}$, $Y=\hat{I}_{src}$ (source view projected to reference view) 
and backward ($\mathcal{L}^{backward}_{SSIM}$) with $X=\hat{I}_{ref}$, $Y=I_{src}$ (reference view projected to source view), 
with no mask $M$ and $K=1$, Eq. \ref{eq:ssim-general-mvs}. The final value of $\mathcal{L}_{SSIM}$ is the 
summation of the forward and the backward components.

As explained earlier, $\mathcal{L}_{SSIM}$ is usually 
combined with a uniformly weighted loss function like $L_1$ and $L_2$, to choose 
the best of both the individual loss terms. The 
combined formulation is shown in Eq. \ref{eq:mix-ssim}. Dai et al. \cite{dai2019mvs2} and Yu et al. \cite{yu2020p} use 
Eq. \ref{eq:mix-ssim} with $\alpha=0.85$ in an MVS framework. This formulation 
finds widespread use in monocular depth estimation problems. Monocular methods like
\cite{wang2020geometric, Yin2018GeoNet}
use Eq. \ref{eq:mix-ssim} with $\alpha = 0.85$ with target and novel view synthesized depth 
maps as $X$ and $Y$. Godard et al. \cite{godard2017unsupervised} use  $\mathcal{L}^{mix}_{SSIM}$ 
(Eq. \ref{eq:mix-ssim}) between the input RGB image and reconstructed new view RGB image with $\alpha=0.85$
to enforce structural similarity. Zhao et al. \cite{zhao2019geometry} with symmetric domain adaptation, 
real-to-synthetic and synthetic-to-real, apply Eq. \ref{eq:mix-ssim} both ways to enforce 
structural similarity during the transition from one domain to the other.

Chen et al. \cite{chen2019self} use  $\mathcal{L}^{mix}_{SSIM}$, Eq. \ref{eq:mix-ssim}, with 
a slight modification to make it an adaptive loss in optical flow estimation. For scene
structures that can not be explained by global rigid motion, it adapts to a more flexible 
optical flow estimation by updating the channel parameters only for the configurations 
that closely explain the displacement. It is represented as the minimum 
error between the optical flow and the rigid motion displacements, 

\begin{equation}
    L^{adaptive}_{SSIM} = min \{ \mathcal{L}^{mix}_{SSIM_{flow}}, \mathcal{L}^{mix}_{SSIM_{rigid}}\}
\end{equation}

\noindent and it uses $\alpha=0.85$ in Eq. \ref{eq:mix-ssim}. We only discuss the $\mathcal{L}^{mix}_{SSIM}$ formulation 
here and refer to \cite{chen2019self} for other optical flow related details.

%%%%%%%%%%%%%%%%%%%%%%%%%%%%%%%%%%%%%%%%%%%%%%%%%%%%%%%%%%%%%%%%%%%%%%%%%%%%%%%%%%%%%%%%%%%%%%%%%%%%%%%%%%%%%%%%%%%
\subsection{Edge-Aware Smoothness Constraint}\label{subsec:smooth}

% Why is it required, its role, 
% 1) first-order and its role, 2) second-order and its role. 
% 3) combination of first and second order
% sometimes used as regularization terms. 

The smoothness constraint finds its origin in the optical flow estimation problem. 
It was first applied by Uras et al. \cite{uras1988computational} to estimate 
consistent optical flow from two images. Brox et al. \cite{brox2004high} 
further explained the concept under three assumptions for the optical flow framework, 
i.e. the gray value constancy assumption, the gradient constancy assumption, and 
the smoothness assumption. 

Since the beginning of the optical flow estimation 
problem, it has been assumed that the gray value of a pixel does not 
change on displacement under a given lighting condition \cite{brox2004high}. 
But brightness changes in natural scenes all the time. Therefore, it is 
useful to allow small variations in gray values but find a different criterion that 
remains relatively invariant under gray value changes, i.e. gradient 
constancy under displacement. 
This brought about the third assumption; while discontinuities are assumed to be present 
at the boundaries of the object in the scene, piecewise smoothness can 
be assumed in the flow field \cite{brox2004high}. To achieve this smoothness 
in flow estimation, a penalty on the flow field was applied. 

% \begin{align}
%     \label{eq:gray-value-assumption}
%     I(x,y,t) &= I(x + u, y+v, t+1)\\
%     \label{eq:grad-consistency-assumption}
%     \triangledown I(x,y,t) &= \triangledown I(x+u, y+v, t+1)\\
%     \label{eq:smooth-energy}
%     E_{smooth} (u,v) &= \int \psi (|\triangledown u|^2 + |\triangledown v|^2) dx
% \end{align}

In the optical flow framework, objects are assumed to be moving with a fixed camera. In 
an MVS framework, the objects are assumed to be fixed and the camera moves 
around a fixed point. The relative motion of an object can be viewed 
as a moving camera to pose it as an MVS problem, see Fig. \ref{fig:depth-flow-motion}. 
With this assumption, the smoothness 
constraint can be applied to the depth estimation problem. Initially, only the 
first-order smoothness constraints were used
\cite{eigen2015predicting, garg2016unsupervised, godard2017unsupervised, Yin2018GeoNet, mahjourian2018unsupervised, yu2020p}. 
Yang et al. \cite{yang2017unsupervised} introduced a 
second-order smoothness constraint for regularization along with 
the first-order smoothness term. Subsequently, both the first and the second-order smoothness 
terms have been used in many works 
\cite{dai2019mvs2, huang2021m3vsnet, dong2022patchmvsnet, yang2023geometry}.
In the depth estimation framework, smoothness constraint is applied 
between the gradient of input images $(I)$ and the estimated depth maps $(\hat{D})$,

\begin{align}
    \label{eq:first-order-smooth}
    \mathcal{L}_{smooth_{\triangledown}} &= \sum || \partial_x \hat{D}||. e^{-||\partial_x I ||} + || \partial_y \hat{D}||. e^{-||\partial_y I ||}\\
    \label{eq:second-order-smooth}
    \mathcal{L}_{smooth_{\triangledown^2}} &= \sum || \partial^2_x \hat{D}||. e^{-||\partial^2_x I ||} + || \partial^2_y \hat{D}||. e^{-||\partial^2_y I ||}\\
    \label{eq:first-second-smooth}
    \mathcal{L}_{smooth} &= \alpha . \mathcal{L}_{smooth_{\triangledown}} + \beta . \mathcal{L}_{smooth_{\triangledown^2}} ,
\end{align}

\noindent where $(\mathcal{L}_{smooth_{\triangledown}})$ and second $(\mathcal{L}_{smooth_{\triangledown^2}})$ 
represents the first and second-order smoothness constraints. $\mathcal{L}_{smooth}$ shows the combined 
formulation with $\alpha >0$ and $\beta > 0$ as a scaling factor.

Garg et al. \cite{garg2016unsupervised} use the first order formulation with $L_2$ 
norm as a regularization term to achieve smoothness in estimation. Mahjourian et al. 
\cite{mahjourian2018unsupervised} use the first-order formulation for monocular video 
depth estimation. Most self-supervised/unsupervised MVS frameworks 
 use first-order smoothness constraints \cite{li2022ds, yu2020p, liu2023geometric, khot2019learning, mallick2020learning}.
Zhao et al. \cite{zhao2019geometry} with its symmetric domain adaptation for 
monocular depth estimation uses first-order constraint in both domains. Other 
monocular depth estimation methods 
\cite{godard2017unsupervised, Yin2018GeoNet, hu2019revisiting, wang2020geometric}
also apply the first-order smoothness constraint 
as defined in Eq. \ref{eq:first-order-smooth}.

Yang et al. \cite{yang2017unsupervised} only uses second order formulation, 
Eq. \ref{eq:second-order-smooth}, as a regularization term in monocular video-depth
estimation framework. Learning from it, more recent MVS frameworks combine 
both the first-order and the second-order formulations   
\cite{huang2021m3vsnet, dai2019mvs2, dong2022patchmvsnet},
as shown in Eq. \ref{eq:first-second-smooth}. Inspired from 
\cite{godard2017unsupervised} and \cite{hu2019revisiting}, Yang et al. 
\cite{yang2023geometry} apply the combined edge-aware smoothness constraint, Eq. \ref{eq:first-second-smooth},
 in monocular endoscopy. All of these methods use $\alpha = \beta = 1$ in Eq. \ref{eq:first-second-smooth}.

%%%%%%%%%%%%%%%%%%%%%%%%%%%%%%%%%%%%%%%%%%%%%%%%%%%%%%%%%%%%%%%%%%%%%%%%%%%%%%%%%%%%%%%%%%%%%%%%%%%%%%%%%%%%%%%%%%%
\subsection{Consistency Regularization}\label{subsec:regularization}

Deep learning-based frameworks inherently suffer from over-parameterization problems. 
One of the most efficient methods to counter it is to regularize the loss function. 
Photometric consistency, Sec. \ref{subsec:photometric-consistency}, which enforces 
geometrical consistency at the pixel level, is highly susceptible to change in lighting 
conditions. Many MVS methods employ different consistency regularization techniques
to effectively handle this problem \cite{yang2017unsupervised, garg2016unsupervised, xu2022semi}.

As discussed in Sec \ref{subsec:smooth}, first-order and second-order gradients 
are often used for this task \cite{yang2017unsupervised, garg2016unsupervised}. 
Garg et al. \cite{garg2016unsupervised} argue that the photometric loss is 
non-informative in homogeneous regions of a scene, which leads to multiple warps generating 
similar disparity outcomes. They use $L_2$ regularization 
($\mathcal{L}_{\triangledown} = || \triangledown \hat{D}||^2$, where D is the disparity map) 
on the disparity discontinuities as a prior, 
and they also recommend using other robust penalty 
functions \cite{brox2004high, zach2007duality} as an alternative regularization term.
Yang et al. \cite{yang2017unsupervised} use a spatial smoothness penalty with $L_1$
norm of the second-order gradient of depth, 
$\mathcal{L}_{\triangledown^2} = \sum_{d \in x,y} |\triangledown^2_d \hat{D}|.e^{-\alpha|\triangledown I|}; \alpha > 0$. They 
encourage depth values to align in the planar surface when no image gradient appear. 

Xu et al. \cite{xu2022semi} apply consistency regularization in the semi-supervised MVS 
method. The proposed regularization minimizes the Kullback-Leibler (KL) divergence 
between the predicted distributions of augmented $(\hat{PV})$ and non-augmented $(PV)$ samples. 
With the $K$ depth hypothesis, the probability volume $PV$, of size $H \times W \times K$, 
is separated into $K$ categories of $HW$ logits. The KL regularize formulation, 

\begin{align}
    \label{eq:KL-regularization}
    \mathcal{L}_{KL} = \frac{1}{HW} \sum^{HW}_{i=1} KL(PV_{p_i} , \hat{PV}_{p_i}),
\end{align}

\noindent where $p_i$ represents a pixel coordinate.

%%%%%%%%%%%%%%%%%%%%%%%%%%%%%%%%%%%%%%%%%%%%%%%%%%%%%%%%%%%%%%%%%%%%%%%%%%%%%%%%%%%%%%%%%%%%%%%%%%%%%%%%%%%%%%%%%%%
\subsection{Structural Consistency in 3D Space}

\begin{figure*}[t]
\begin{center}
    \includegraphics[width=0.95\textwidth]{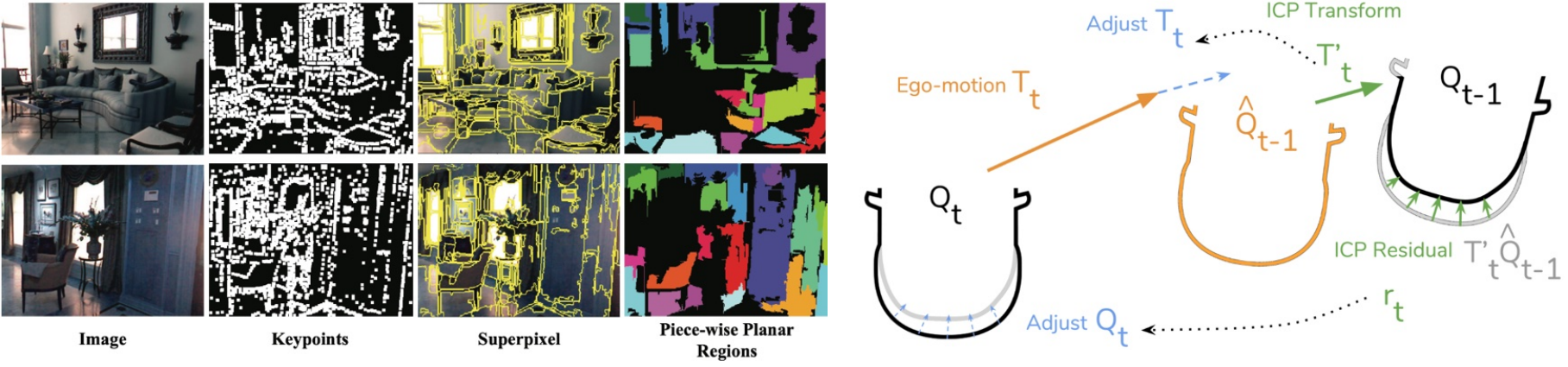}
    \caption{Left: Sample input images, key points, superpixels, and 
    piece-wise planar regions obtained from large superpixels (Fig. reprinted from \cite{yu2020p}).
    Right: Point cloud matching process and approximate gradients for the top view of 
    a car, Fig. reprinted from \cite{mahjourian2018unsupervised}.}
    \label{fig:palanr-and-3d-alignment}
\end{center}
\end{figure*}

Structural consistency is not limited to the 2D image plane, and can easily be 
extended to camera 3D space (a virtual 3D space after applying extrinsic 
camera parameters to a scene) or 3D point clouds (a reconstruction of points in a 3D space). In this section, 
we discuss two such methods \cite{yu2020p,mahjourian2018unsupervised}
that use structural consistency in 
3D space alongside other geometric constraints in an end-to-end framework. 

\subsubsection{Planar Consistency:}

Planar consistency \cite{yu2020p} is based on the assumption that most 
homogeneous-color regions in an indoor scene are planar regions and have continuous depth. 
Extraction of such piece-wise planar regions is a three-step process. 
Given an input image $I$, the key points are first extracted. The key points in the input image 
are then used to extract superpixels. Finally, a segmentation algorithm is used to apply 
a greedy approach to segment areas with low gradients to produce more planar regions. There 
are many ways to extract key points and superpixels, \cite{yu2020p} uses 
Direct Sparse Odometry \cite{engel2018DSO} to extract key points and  Felzenszwalb superpixel 
segmentation \cite{felzenszwalb2004efficient} for superpixels' and planar region segmentation. 

The left side images in Fig. \ref{fig:palanr-and-3d-alignment} show the steps of obtaining planar 
regions in two indoor scenes. 
For an image $I$, after extracting superpixels, a threshold is applied to only keep 
regions with greater than $1000$ pixels. It is assumed that most planar regions occupy 
large pixel areas. With the extracted super pixels $SPP_m$ and corresponding depth $D(p_n)$, 
we first back project all points $p_n$ into 3D space ($p^{3D}_{n}$), $p^{3D}_{n} = D(p_n) K^{-1} p_n, p_n \subseteq SPP_m$. 
Using the plane parameter $A_m$ of $SPP_m$, the plane is defined in 3D space, $A^{T}_{m} p^{3D}_{n} = 1$.
$A_m$ is calculated using two matrices, $Y_m = [1,...,1]^T$ and $P_n = [p^{3D}_{1}, ..., p^{3D}_{n}]$,
$P_n A_m = Y_m; A_m = (P^T_n P_n + \epsilon E)^{-1} P^T_n Y_m$, where $E$ is an identity matrix and $\epsilon$ is a for 
numerical stability. With planar parameters, a fitted planar depth for each pixel in all superpixels 
can be retrieved to estimate the planar loss,

\begin{align}
    \label{eq:fitted-depth}
    D'(p_n) &= (A^T_m K^{-1} p_n)^{-1}\\
    \label{eq:planar-loss}
    \mathcal{L}_{planar} &= \sum^{M}_{m=1} \sum^{N}_{n=1} |D(p_n) - D'(p_n)| ,
\end{align}

\subsubsection{Point Cloud Alignment:}

Mahjourian et al. \cite{mahjourian2018unsupervised} use another approach to 
align the 3D point clouds of two consecutive frames ($Q_{t-1}, Q_t$) in video depth estimation. 
They directly compare the estimated point cloud associated with respective frames ($\hat{Q}_{t-1}$ and $\hat{Q}_{t}$), 
using a well-known rigid registration method,
Iterative Closest Point (ICP) \cite{besl1992method, chen1991object, Rusinkiewicz2001efficient}, which  
computes a transformation to minimize the point-to-point distance between two point clouds. They 
alternate between computing correspondences between 3D points and best-fit transformations between 
the two point clouds. For each iteration, they recompute the correspondence with the previous iteration's 
transformation applied. 

ICP is not differentiable, but its gradients can be approximated using the 
product it computes as part of the algorithm, allowing back-propagation. 
It takes two point clouds $A$ and $B$ as input and produces two outputs. 
First, is the best-fit transformation $T'$ which minimizes the distance between 
the transformed points in $A$ and $B$,
and second is the residual $r^{ij}$, $r^{ij} = A^{ij} - T'^{-1}.B^{c(ij)}$, 
which reflects the residual distances between corresponding points after ICP's minimization. 
The alignment loss is given as,  

\begin{align}
    \label{eq:3d-alighment-loss}
    \mathcal{L}_{3DAlignment} = || T'  - I||_1 + ||r||_1; I= 1 ,
\end{align}

\noindent where $I$ is the identity matrix.

For each frame $t$ of the video, if the alignment of the estimate is not perfect, 
the ICP algorithm produces a transformation $T'_t$ and $r_t$, which can be used to 
adjust the estimates towards initial alignment \cite{mahjourian2018unsupervised}. 
Right side of 
Fig. \ref{fig:palanr-and-3d-alignment} shows the steps of alignment loss.

%%%%%%%%%%%%%%%%%%%%%%%%%%%%%%%%%%%%%%%%%%%%%%%%%%%%%%%%%%%%%%%%%%%%%%%%%%%%%%%%%%%%%%%%%%%%%%%%%%%%%%%%%%%%%%%%%%%

\section{Normal-Depth Orthogonal Constraint}
\label{sec:surface-normal}
%%%%%%%%%%% list of methods %%%%%%%%%
%% 1. Surface normal
%% 2. Depth-Normal constraints
%% 3. 
%% 4. 

\begin{figure*}[t]
\begin{center}
    \includegraphics[width=0.9\textwidth]{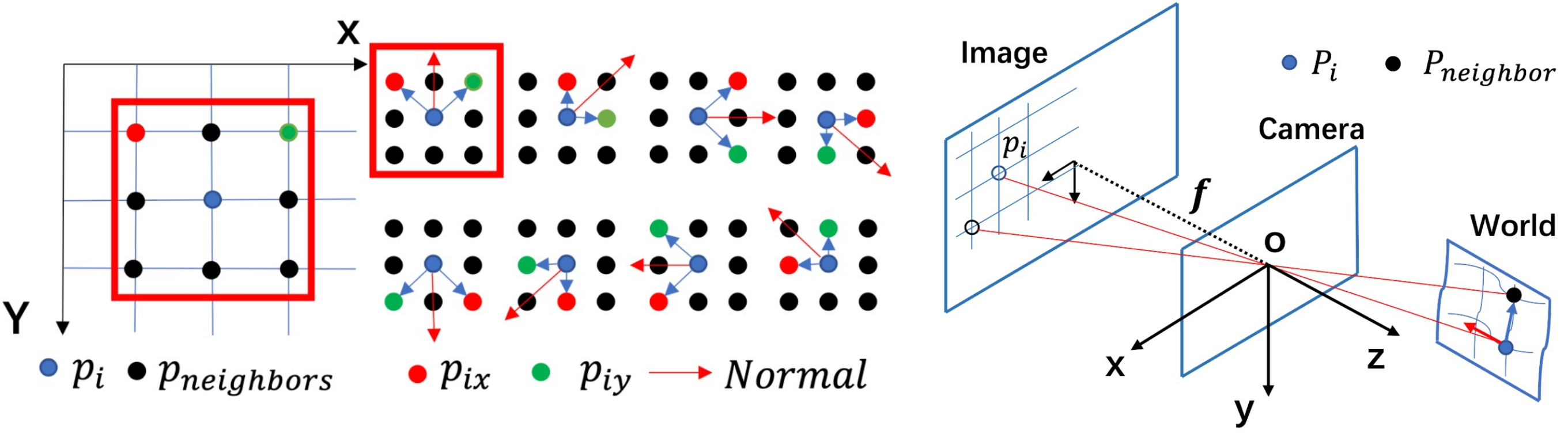}
    \caption{Left: estimation of normal from depth. Right: estimation of 
    depth from normal. Figure reprinted from \cite{huang2021m3vsnet}}
    \label{fig:normal-depth-constraints}
\end{center}
\end{figure*}

%%%%%%%%%%%%%%%%%%%%%%%%%%%%%%%%%%%%%%%%%%%%%%%%%%%%%%%%%%%%%%%%%%%%%%%%%%%%%%%%%%%%%%%%%%%%%%%%%%%%%%%%%%%%%%%%%%%

Surface normals are important 'local' features of the 3D point cloud of 
a scene, which can provide promising 3D geometric cues to 
estimate geometrically consistent depth maps. In the 3D world coordinate 
system, the vector connecting two pixels in the same plane should be
orthogonal to their direction of normal. Enforcing normal-depth 
orthogonal constraints tend to improve depth estimates in 
3D space \cite{wang2016surge, eigen2015predicting, yang2017unsupervised}. 
Application of these constraints during the learning process helps model 
better understand the representations related to cluttered backgrounds, 
repetitive patterns and texture-less regions. These can efficiently be 
utilized as part of an objective function in a deep learning framework.

\paragraph{Depth to normal:}
Given depth $D_i$, 
to estimate the normal of each central pixel $p_i$, we need 
to consider the neighboring pixels $p_{neighbors}$. 
Fig. \ref{fig:normal-depth-constraints} (left) shows $8$ 
neighbor convention to compute the normal of the central pixel. 
Two neighbors, $p_{ix}$ and $p_{iy}$ are selected from $p_{neighbors}$
for each central pixel $p_i$, with depth value as $D_i$ 
and camera intrinsics $K$, to estimate normal $\tilde{N_i}$. 
First, we project the depth in 3D space, and then take the cross
product between vectors $\overrightarrow{P_i P_{ix}}$ and 
$\overrightarrow{P_i P_{iy}}$ to estimate the normal,

\begin{align}
    \label{eq:pixel-projection}
    P_i &= K^{-1} D_i p_i\\
    \label{eq:each-normal}
    \tilde{N_i} &= \overrightarrow{P_i P_{ix}} \times \overrightarrow{P_i P_{iy}}   
\end{align}

\noindent With $8$ such estimates of normal $\tilde{N_i}$ for
each central pixel, we estimate the final normal $N_i$ as the mean value of all the estimates using 
$N_i = \frac{1}{8} \sum^{8}_{i=1} (\tilde{N_i})$.

\paragraph{Normal to depth:}
Many methods \cite{huang2021m3vsnet, yang2017unsupervised, yin2019enforcing}
use normal to depth estimation to refine the depth values $D_i$
using the orthogonal relationship. For each pixel $p_i(x_i, y_i)$,
the depth of each neighbor $p_{neighbor}$ should be refined. 
The corresponding 3D points are $P_i$ and $P_{neighbor}$, and
the central pixel $P_i$'s 
normal $\overrightarrow{N_i}(n_x, n_y, n_z)$. The depth of 
$P_i$ is $D_i$ and $P_{neighbor}$ is $D_{neighbor}$. Using the 
orthogonal relations 
$\overrightarrow{N_i} \perp \overrightarrow{P_i P_{neighbor}}$, we 
can write,

\begin{align}
    [K^{-1} D_i p_i - K^{-1} D_{neighbor} P_{neighbor}]\begin{bmatrix} 
        n_x\\
        n_y\\
        n_z\\
        \end{bmatrix} = 0
\end{align}

\noindent With depth 
estimates coming from eight neighbors, we need a method for 
weighting these values to incorporate discontinuity of normal in some edges
or irregular surfaces. Generally, the weight $w_i$ is inferred from 
the reference image $I_i$, making depth more 
geometric consistent. The value of $w_i$ depends on the gradient 
between $p_i$ and $p_{neighbor}$. 
The larger gradients 
represent less reliable refined depth. Given the eight neighbors, the final refined depth 
$\tilde{D}_{neighbor}$  is a weighted sum of eight different 
directions as \cite{huang2021m3vsnet},

\begin{align}
    \label{eq:weight-depth-neighbor}
    \tilde{D}_{neighbor} = \sum^8_{i=1} \frac{w_i}{\sum^8_{i=1} w_i} D_{neighbor}
\end{align}

\noindent where $    w_i = e^{- \alpha |\triangledown I_i|}$.
This is the outcome of the regularization in 3D 
space, improving the accuracy and continuity of the estimated 
depth maps.

Yang et al. \cite{yang2017unsupervised} use a similar formulation to enforce geometric consistency in 
unsupervised video depth estimation problems.
Wang et al. \cite{wang2016surge} use the orthogonal compatibility principle to bring 
consistency in the normal directions of two pixels falling on the same plane.
Eigen and Fergus \cite{eigen2015predicting} use a single multiscale CNN
to estimate depth, surface normal, and semantic labeling. For surface 
normal estimation at each pixel, they predict the $x,y$ and $z$
components for each pixel \cite{silberman2012indoor}
and employ elementwise loss 
comparison with dot product,

\begin{equation}
    \mathcal{L}_{normal} = - \frac{1}{N} \sum_i \hat{n}_i \odot n_i ,
    \label{eq:dot-prod-normal}
\end{equation}

\noindent where $N$ is the valid pixels, $n$ and $\hat{n}$ are ground truth and 
the predicted normal at each pixel.

Qi et al. \cite{qi2018geonet} also employ depth-to-normal and 
normal-to-depth networks to regularize the depth estimate in 
3D space following \cite{fouhey2013data}, but they do not 
use 8-neighbor-based calculation. Instead, 
they use a distance-based selection of neighboring 
pixels, 

\begin{equation}
    N_i = \{(x_j, y_j, z_j) || u_i - u_j| < \beta, |v_i - v_j| < \beta, |z_i - z_j| < \gamma z_i\}
\end{equation}

\noindent where $u_i, v_i$ are the 2D coordinates, $(x_i, y_i, z_i)$ are the 3D 
coordinates, and $\beta$ and $\gamma$ are hyper-parameters controlling the 
size of the neighborhood along $x-y$ and depth axes respectively. They 
use $L_2$ norm between the ground truth normal and the estimated normal 
as the loss $\mathcal{L}_{normal}$ in end-to-end deep learning framework.

Hu et al. \cite{hu2019revisiting} use ground truth and estimated depth
map gradients to measure the angle between their surface normals using
$n^{\hat{d}}_{i} = [- \triangledown_x \hat{d}_i, - \triangledown_y \hat{d}_i, 1]^T$ and 
$n^{d}_{i} = [- \triangledown_x d_i, - \triangledown_y d_i, 1]^T$.
The loss, $\mathcal{L}_{normal}$
is estimated using, 

\begin{align}
    \label{eq:grad-based-normal-loss}
    \mathcal{L}_{normal} &= \frac{1}{N} \sum^{N}_{i=1} \left(1 - \frac{\langle n^{\hat{d}}_{i}, n^d_i \rangle}{\sqrt{\langle n^{\hat{d}}_{i},n^{\hat{d}}_{i} \rangle}  \sqrt{\langle n^{d}_{i},n^{d}_{i} \rangle}} \right)
\end{align}

\noindent where 
$\langle.,.\rangle$ denotes the inner product of vectors. This loss is 
sensitive to small depth structures \cite{hu2019revisiting}. 
Yang et al. \cite{yang2023geometry} use the same method of estimating normal 
for monocular depth estimation problems in endoscopy applications.

Advantages of estimating normals in a depth estimation 
frameworks are that it provides an explicit understanding of normal during the 
learning process, higher-order interaction between the 
estimated depth and ground truth, and flexibility to 
integrate additional operations over normal
\cite{yang2017unsupervised}. On the other hand, it is sensitive 
to noise in the ground truth depth maps and the estimated depth
maps, and only considers local information which may 
not align with the global structure. 

To enforce robust higher-order geometric supervision in 3D space, 
Yin et al. \cite{yin2019enforcing} propose virtual normal (VN) estimation. 
VN can establish 3D geometric connections between regions in a much 
larger range. To estimate VN, $N$ group points from the depth 
map, with three points in each group, are sampled. The selected point 
has to be non-colinear. The three points establish a plane and the 
corresponding normal is estimated. Similarly, ground truth normal $(n_i)$ 
are estimated and compared with the normals corresponding to the 
estimated depth maps $(\hat{n}_i)$, 

\begin{equation}
    \mathcal{L}_{Normal_{robust}} = \frac{1}{N}\left ( \sum^{N}_{i=1} || \hat{n}_{} - n_{i}||_1  \right) ,
    \label{eq:virtual-normal}
\end{equation}

\noindent Naderi et al. \cite{naderi2022monocular} use a similar 
formulation in a monocular depth estimation problem to enforce 
higher-order robust geometric constraints for depth estimation. 

\begin{figure*}[t]
\begin{center}
    \includegraphics[width=0.5\textwidth]{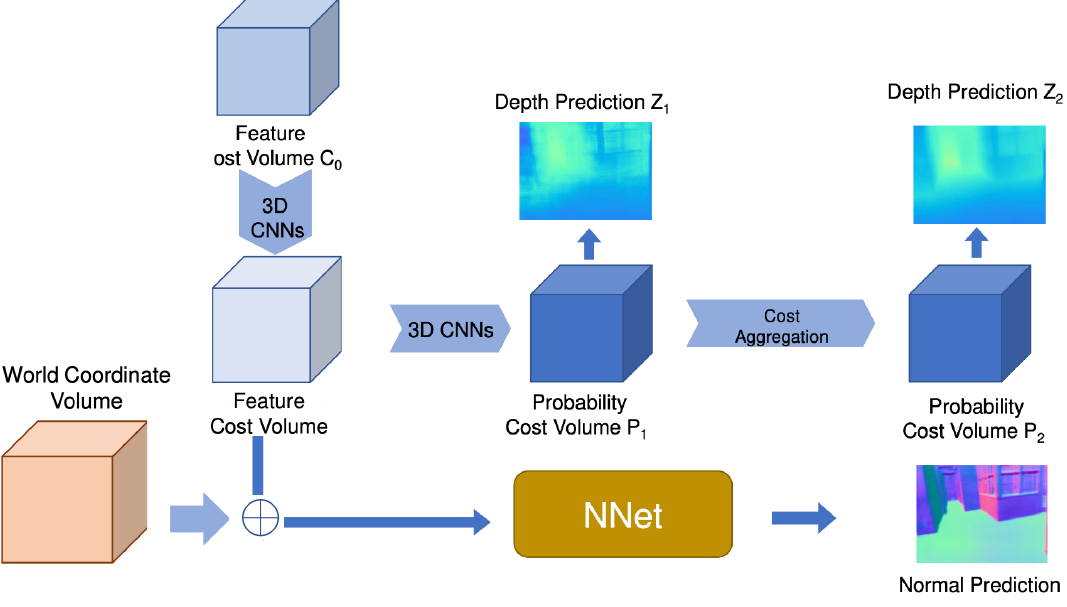}
    \caption{Normal-depth joint learning approached. Depth and normal are estimated using the same feature cost volume \cite{kusupati2020normal}. The above diagram is a slight modification from 
    Kusupati et al. \cite{kusupati2020normal} to only show the joint estimation setup.}
    \label{fig:depth-normal-joint}
\end{center}
\end{figure*}

\paragraph{Normal-depth joint learning approach:}

Kusupati et al. \cite{kusupati2020normal} develop a normal-assisted depth estimation
algorithm that couples the learning of multi-view normal estimation and multi-view depth
estimation process. It uses a feature cost volume to estimate the depth map and 
the normal. A cost volume provides better structural representation to facilitate 
better learning on the image features. 
Specifically, Kusupati et el. \cite{kusupati2020normal} estimate two depth maps
$Z_1$ and $Z_2$ using 3D-CNNs, as shown in Fig. \ref{fig:depth-normal-joint}, and uses 
a 7-layered CNN (NNet, Fig. \ref{fig:depth-normal-joint}) to estimate the normal. The world coordinate volume is concatenated 
with the initial feature volume to provide a feature slice to NNet as input. 
NNet predicts normals for each feature slice, which are later averaged to get the final 
normal map.

\section{Attention Meets Geometry}
\label{sec:geometric-attention}

Recent advancements in attention-based deep learning frameworks 
have revolutionized the computer vision field. Network 
architecture like a Transformer with a self-attention mechanism \cite{dosovitskiy2020image} 
has become the best-performing network for various vision tasks. 
Its ability to model long-range global-contextual 
representation by dynamically shifting its attention within the image makes it unique. 
The inputs to the attention module are usually named 
query (Q), key (K), and value (V). Q retrieves information 
from V based on the attention weights, $Attention(Q,K,V) = \mathcal{A}(Q, K)V$ where $\mathcal{A}$ 
is an attention estimating function, 
where $\mathcal{A}(.)$ is a function that produces similarity scores 
as attention weights between feature embeddings for aggregation. 

The performance of stereo or MVS depth estimation methods depends
on finding dense correspondences 
between reference and source images. Recently, Sun et al. \cite{sun2021loftr} 
showed that features extracted using a transformer model with self- and cross-attention 
can produce significantly improved correspondences as compared to features extracted using 
convolutional neural networks. These attention mechanisms are designed to  
pay attention to the contextual information and not to geometry-based 
information. Recently, a few methods have modified these attention 
mechanisms to consider geometric information while calculating the 
attention weight \cite{ruhkamp2021attention, naderi2022monocular, zhu2021multi, guo2020normalized}.
In this section, we discuss such methods and their approach to include 
geometric information in attention.

Ruhkamp et al. \cite{ruhkamp2021attention} use geometry to guide spatial-temporal 
attention to guide self-supervised monocular depth estimation method from videos. 
They propose a spatial-attention layer with 3D spatial awareness by exploiting 
a coarse predicted initial depth estimate. With known intrinsic camera parameter K 
and pair of coordinates ($C_i, C_j$) along with their depth estimates $(d_i, d_j)$, 
they back-project the depth values into 3D space using Eq. \ref{eq:back-project}. 
The 3D space-aware spatial attention is then calculated as,

\begin{align}
    \label{eq:back-project}
    p_i = K^{-1}(d_i . C_i); p_j &= K^{-1}(d_j . C_j) ,\\
    \label{eq:spatial-atten-1}
    \mathcal{A}^{spatial}_{i,j} &= exp\left( - \frac{||P_i - P_j||_2}{\sigma}\right) ,\\
    \label{eq:temporal-atten-1}
    \mathcal{A}^{temporal}_{i,j} &= Softmax_j(F^{q^T}_{i} F^{k}_{j}) ,
\end{align}

\noindent where $P_i, P_j$ are treated as K and Q, respectively. They use Eq. \ref{eq:temporal-atten-1}
to estimate the temporal attention for aggregation. The unique formulation of 
the spatial-temporal attention model can explicitly correlate geometrically 
meaningful and spatially coherent features for dense correspondence \cite{ruhkamp2021attention}.

\begin{figure*}[t]
    \centering
    \includegraphics[width=0.5\textwidth]{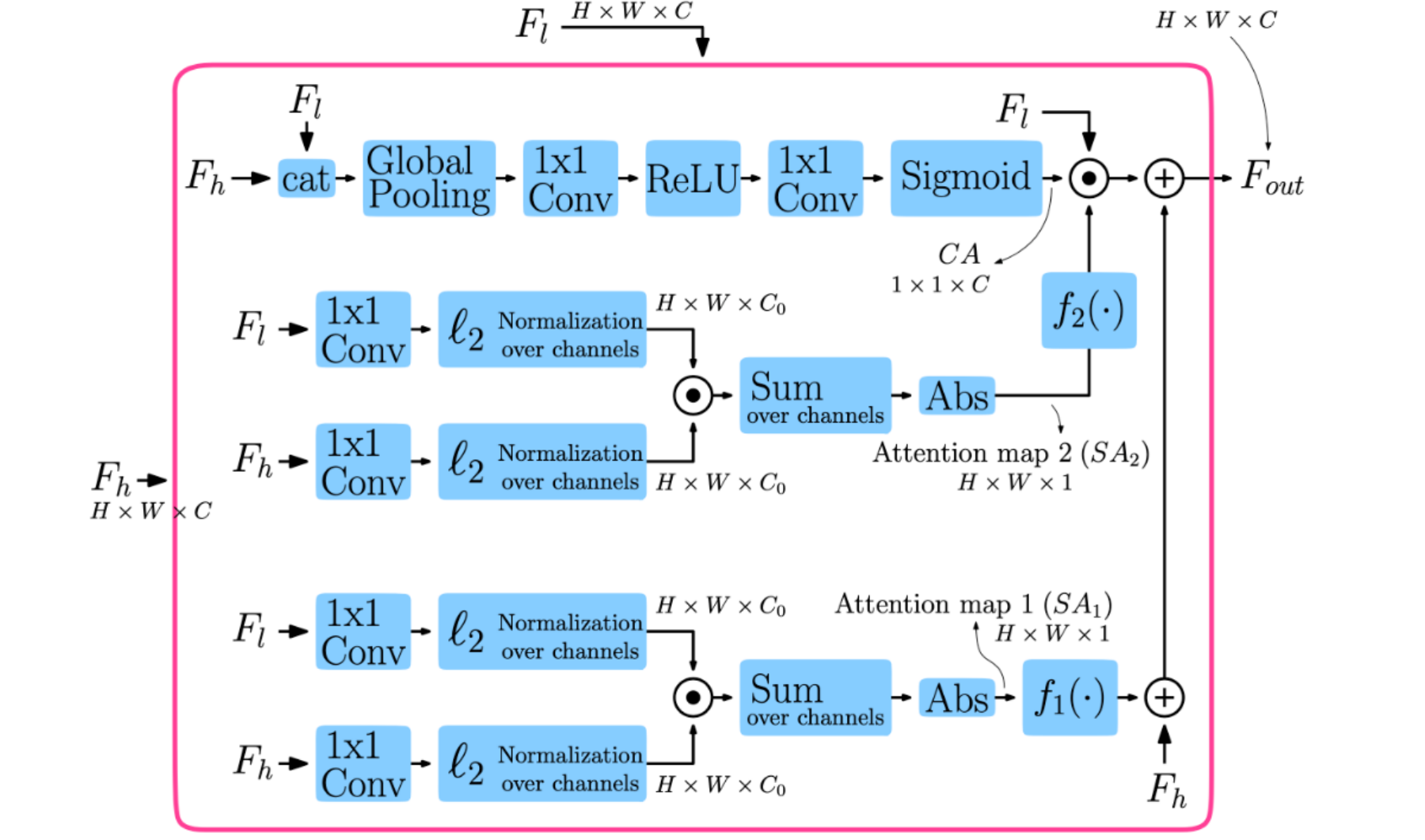}
    \caption{Adaptive Geometric Attention. Figure reprinted from \cite{naderi2022monocular}}
    \label{fig:mono-geo-attention}
\end{figure*}

Naderi et al. \cite{naderi2022monocular} propose adaptive geometric attention (AGA) 
for monocular depth estimation with an encoder-decoder architecture. They apply the AGA module in the 
decoding step utilizing both low-level $(F_L)$ and high-level $(F_h)$ features. Fig. 
\ref{fig:mono-geo-attention} shows the process of calculating AGA. The first row of operations 
in Fig. \ref{fig:mono-geo-attention} shows the steps to calculate channel-attention 
$(\mathcal{CA})$, which produces a $1\times1\times C$ shape attention map and is multiplied 
with $F_l$. The other two rows show two spatial attention $(\mathcal{SA})$
calculations using $\mathcal{SA}_i = |Cosine_{similarity}(E_{l,i}, E_{h,i})|, i = 1,2$. 
The final aggregated 
feature output $(F_out)$ is estimated using $F_{out} = [f_1(\mathcal{SA}_1) + f_2(\mathcal{SA}_2) \times \mathcal{CA}] \times F_l + F_h$, 
where $\mathcal{SA}_1$ is added and $\mathcal{SA}_2$ is multiplied with $F_L$. $f_1(.)$ and $f_2(.)$ 
are introduced to enhance the sensitivity to any non-zero correlation between $F_l$ and $F_h$, 
$f(\mathcal{SA}) = \mathcal{SA}$ is with no-enhanced sensitivity, whereas $f(\mathcal{SA}) = \mathcal{SA} . \, exp(\mathcal{SA})$ 
shows the formulation of the enhanced sensitivity attention map. 

Zhu et al. \cite{zhu2021multi} use two types of transformer modules to extract 
geometry-aware features in an MVS pipeline, a global-context transformer module 
and a 3D geometry transformer module.
The global-context transformer module extracts 3D-consistent reference features $(T_r)$
which are then used as input to the 3D-geometry transformer module to facilitate cross-view attention. 
$(T_r)$ is used to generate K and V to 
enhance interaction between reference and source view features for obtaining 
dense correspondence.

Guo et al. \cite{guo2020normalized} use a geometry-aware attention mechanism 
for image captioning. Unlike other methods discussed above, 
they do not modify the attention mechanism 
itself but add a bias term which makes the feature extraction biased towards
specific content.  Their attention module is similar to \cite{dosovitskiy2020image} 
apart from the added bias term in score $E$ calculation, 
$E = QK^T + \phi(Q',K',G)$, where
$G_{ij}$ is the relative geometry feature between two objects $i$ and $j$. 
There are two terms in score $E$, the \textit{content-based weights} 
and \textit{geometric bias}. They propose three different ways 
of applying geometric bias. 
\textit{content-independent} bias ($\phi^{independent}_{ij} = ReLU(w^T_g G_{ij})$) assumes 
static geometric bias, i.e. same geometric bias is applied to all the K-Q pairs.
The \textit{query-dependent} bias provides geometric information 
based on the type of query ($\phi^{query}_{ij} = Q^{'T}_g G_{ij}$) and \textit{key-dependent} bias provides 
geometric information based on the clues present in keys, $\phi^{key}_{ij} = K^{'T}_g G_{ij}$

\section{Learning Geometric Representations}
\label{sec:geometric-representation}
%%%%%%%%%%% list of methods %%%%%%%%%
Besides using direct methods of enforcing geometry-aware constraints, 
providing geometric guidance, and exploiting orthogonal relations between 
depth and normals, there are indirect ways of learning 
structurally consistent feature representations. For example, features  
with a high level of semantic information are more likely to retain 
the structural consistency of objects compared to features with
a low level of semantic information. The generation of pseudo-labels based on geometric consistency
can be utilized for self-supervised training. Robust feature representation can be 
learned using suitable data-augmentation techniques, and  
attaching semantic segmentation information of objects, like a co-segmentation map can 
improve the structural consistency of objects' features. Contrastive 
learning with positive and negative pairs can guide a model to learn 
better representations that are geometrically consistent features. 
All these constraints helps the model learn a geometry-based representations 
which are able to help with problem like cluttered background, repetitive patterns and texture-less regions. 
These constraints are mostly utilized as part of an objective function in a learning-based framework. 
We discuss all these methods in this section.

%%%%%%%%%%%%%%%%%%%%%%%%%%%%%%%%%%%%%%%%%%%%%%%%%%%%%%%%%%%%%%%%%%%%%%%%%%%%%%%%%%%%%%%%%%%%%%%%%%%%%%%%%%%%%%%%%%%
\subsection{High-Level Feature Alignment Constraints}

In deep learning-based frameworks for depth estimation, the quality of the extracted features
directly impacts the quality of depth estimates. The poor quality of extracted features 
can greatly impact the local as well as global structural patterns. One way to handle this 
problem is by guiding the extracted features with better representations from an 
auxiliary pre-trained network. While the integrated feature extraction network in 
the depth estimation pipeline can learn useful features, it still lags in 
learning higher-level representations compared to a network explicitly designed 
to learn deep high-level representations, like VGG \cite{simonyan2014very}, 
Inception \cite{szegedy2017inception}, ResNet \cite{he2016deep}, etc. 
To enforce the feature alignment constraint, Johnson et al. \cite{johnson2016perceptual}
propose two loss functions, feature reconstruction loss and style reconstruction loss. 

Feature reconstruction loss, $\mathcal{L}_{feature} = ||F_{target} - F_{source}||_{L_i}$, encourages the model to 
generate source features similar 
to the target features at various stages of the network \cite{johnson2016perceptual}. The strong 
feature representations from target features help mitigate confusions arising because of cluttered background and 
repetitive patterns. This also helps the model in extracting good representation in low-texture or texture-less regions \cite{huang2021m3vsnet, dong2022patchmvsnet, zhao2019geometry}.
Minimizing feature reconstruction loss for early layers improves local visual as well as 
structural patterns, while minimizing it for higher layers improves overall structural 
patterns \cite{johnson2016perceptual}. Feature reconstruction loss fails to preserve 
color and texture, which is handled by style reconstruction loss \cite{johnson2016perceptual}.

The generalized formulation of feature reconstruction loss,
where $F_{target}$ is the target feature, $F_{source}$ is the source feature and $L_i$ denotes $L_1$ or $L_2$ norm. 
A similar formulation is adopted by Huang et al. \cite{huang2021m3vsnet}, dubbed as 
feature-wise loss, in an unsupervised MVS framework. Using a pre-trained VGG-16 network, high-level 
semantically rich features are extracted at layers 8, 15, and 22 for both the reference $(F_{ref})$ and the 
source $(F_{src})$ images. The features from the source images are warped to the reference view $(\hat{F}_{src})$ using 
camera parameters and used in the feature reconstruction loss, 
$\mathcal{L}_{feature} = \frac{1}{N} \sum \left( F_{ref} - \hat{F}_{src} \right)* M_{ref}$. $M$ is 
the reference view mask to handle occlusion and $N$ is the number of source view images. Dong et al. 
\cite{dong2022patchmvsnet} use a similar formulation and extract feature from $8^{th}$ and $15^{th}$ 
layers.

Applying feature alignment loss helps the model extract high-level semantically rich features for 
depth estimation can also be used to generate geometrically consistent and style-conforming
new RGB image, which is then used as input to the depth estimation network. Zhao et al. 
\cite{zhao2019geometry} and Xu et al. \cite{xu2022semi} use such an approach to 
fill generate synthetic input images that are geometrically consistent across 
views and close to the original data distribution. 

Zhao et al. \cite{zhao2019geometry} generate 
synthetic RGB images, and they apply feature and style reconstruction losses
at the final image resolution. They learn bidirectional translators from source to target $G_{s2t}$,
and from target to source $G_{t2s}$, to bridge the gap between the source domain (synthetic) $X_s$ and 
the target domain (real) $X_t$. Specifically, image $x_s$ is sequentially fed to 
$G_{s2t}$ and $G_{t2s}$ generating a reconstruction of $x_s$ and vice-versa for $x_t$. These 
are then compared to the original input,

\begin{equation}
    \mathcal{L}_{Feature_{cycle}} = ||G_{t2s}(G_{s2t}(x_s)) - x_s ||_1 + ||G_{s2t}(G_{t2s}(x_t)) - x_t ||_1 ,
\end{equation}

While Zhao et al. \cite{zhao2019geometry} apply $\mathcal{L}_{feature_{cycle}}$ at the RGB image level, 
Xu et al. \cite{xu2022semi}  use feature and style reconstruction loss in a semi-supervised 
MVS framework. They use a geometry-preserving module
to generate geometry and style-conforming RGB images using a labeled real image.
They employ a geometry-preserving module to generate unlabeled RGB images which are later used 
as input to estimate depth maps. 
The geometry-preserving module includes a spatial propagation network (SPN) with 
two branches, propagation network, and guidance network. 
The labeled image $(I_l)$ is used as input to the guidance network to generate 
an alternate view RGB image $(I_g)$. This RGB image is used as input to the depth estimation 
pipeline to generate a corresponding depth map $D_g$, which is then 
warped to the original view ($\hat{D}_g$) to compare with the ground truth depth map $D_l$.
The loss function is $\mathcal{L}_{feature_{style}} = || D_l - \hat{D}_g  ||^2_2$ \cite{xu2022semi}.
With this setup, they use labeled images to generate geometrically conforming alternate view unlabeled images, and use 
them in the depth estimation pipeline without having to create new labeled data.

\begin{figure*}[t]
    \centering
    \includegraphics[width=0.8\textwidth]{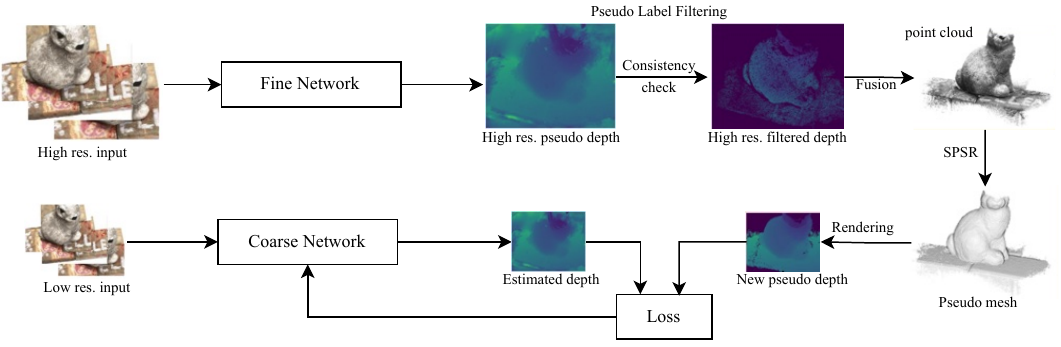}
    \caption{Pseudo-label generation method of \cite{yang2021self}. SPSR is 
    Screened Poisson Surface Reconstruction \cite{kazhdan2013screened}. Figure inspired from \cite{yang2021self}.}
    \label{fig:pseud-label-filtering}
\end{figure*}

%%%%%%%%%%%%%%%%%%%%%%%%%%%%%%%%%%%%%%%%%%%%%%%%%%%%%%%%%%%%%%%%%%%%%%%%%%%%%%%%%%%%%%%%%%%%%%%%%%%%%%%%%%%%%%%%%%%
\subsection{Pseudo-Label Generation with Cross-View Consistency}

In self-supervised MVS frameworks, pseudo-labels are an effective method of applying geometric constraints. 
Generating pseudo-labels requires application of cross-view 
consistency constraints, which encourages the MVS framework to be geometrically consistent 
during training and evaluation \cite{yang2021self, liu2023geometric}. Since the model 
learns with self-supervision, it also helps with the challenging task of collecting 
multi-view ground truth data. In this section, we discuss three methods of generating 
pseudo-labels for self-supervision, labels from high-resolution training images \cite{yang2021self}, 
sparse pseudo-label generation \cite{liu2023geometric} and semi-dense pseudo-labels \cite{liu2023geometric}. 

Yang et al. \cite{yang2021self} apply pseudo-label learning in four steps. First,
they estimate a depth map based on photometric consistency, Sec. \ref{subsec:photometric-consistency}, 
using a coarse network (low-resolution network). With the initial pseudo-label in hand, they apply two step 
iterative self-training to refine these pseudo-labels, see Fig. \ref{fig:pseud-label-filtering}. 
They utilize fine-grained high-resolution networks 
to refine the initial coarse pseudo-labels utilizing more discriminative features from high-resolution images. 
The fine-network estimates high-resolution labels which are then filtered with a cross-view depth consistency check,
Sec. \ref{sec:cross-view}, utilizing depth re-projection error. 
Finally, the high-resolution filtered pseudo-labels from $N$ different views are fused using a 
multi-view fusion method. It generates a more complete point cloud of the scene. The point cloud is then rendered
to generate cross-view, geometrically consistent new pseudo-labels to guide the coarse network depth 
estimation pipeline. 

Liu et al. \cite{liu2023geometric} use two geometric prior-guided pseudo-label generation methods, 
sparse and semi-dense pseudo-label. For sparse label generation, they use a pre-trained 
Structure from Motion framework (SfM) \cite{schonberger2016structure} to generate a sparse point cloud. 
This sparse point cloud is then projected to generate sparse depth pseudo-labels. Since the sparse point
cloud can provide very limited supervision, they use a traditional MVS framework such as 
COLMAP \cite{schonberger2016pixelwise} with 
geometric and photometric consistency to estimate preliminary depth maps. 
The preliminary depth map then undergoes cross-view 
geometric consistency check to eliminate outliers. This filtered depth map 
is then used as a final pseudo-label for learning.

%%%%%%%%%%%%%%%%%%%%%%%%%%%%%%%%%%%%%%%%%%%%%%%%%%%%%%%%%%%%%%%%%%%%%%%%%%%%%%%%%%%%%%%%%%%%%%%%%%%%%%%%%%%%%%%%%%%
\subsection{Data-Augmentation for Geometric Robustness}

Deep-learning frameworks can always do better with more data \cite{shorten2019survey}, but 
collecting data for stereo or MVS is a difficult task. Applying data augmentation to these 
frameworks makes sense but is not as easy to implement. 
The natural color fluctuation, occlusion, and geometric distortions
in augmented images disturbs the color constancy of images, affecting the
effective feature matching and hence the performance of the 
whole depth estimation pipeline \cite{xu2021self}. Because of these limitations, 
it has seldom been applied 
in either supervised \cite{ding2022transmvsnet, yao2018mvsnet, gu2020cascade} or 
unsupervised \cite{khot2019learning, dai2019mvs2, huang2021m3vsnet} MVS methods.

Keeping these limitations in mind, Garg et al. 
\cite{garg2016unsupervised} use three data-augmentation techniques in an unsupervised stereo depth 
estimation problem. They use \textit{color change} -- scalar multiplication of color channels by a factor $c \in [0.9, 1.1]$, 
\textit{scale and crop} -- the input image is scaled by $2\times$ factor and then randomly cropped 
to match the original input size, and \textit{left-right flip} -- wherein the left and right images are 
flipped horizontally and swapped to get a new training pair. These three simple augmentations lead to
an $8$-fold increase in data and improved localization of object edges in the depth estimates. 

Xu et al. \cite{xu2021self} propose using data augmentation as a regularization technique. 
Instead of optimizing the regular loss function with ground truth depth maps, 
they propose \textit{data-augmentation consistency loss}
 by contrasting data augmentation depth estimates with non-augmented depth estimates. 
 Specifically, given the non-augmented input images $I_{non-aug}$ and 
 augmented input images $I_{aug}$ of the same view, the difference between the estimated
 augmented ($\hat{D}_{aug}$) and non-augmented ($\hat{D}_{non-aug}$) depth maps 
 are minimized \cite{xu2021self}, 
 
\begin{equation}
    \mathcal{L}_{augmentation} = \frac{1}{M} \sum ||\hat{D}_{aug} - \hat{D}_{non-aug}||_2 \odot M_{non-aug}
    \label{eq:data-aug-loss}
\end{equation}
 
\noindent where $M_{non-aug}$ denotes an unoccluded mask under data-augmentation transformation. This 
formulation also helped with cluttered backgrounds and texture-less regions. Xu et al. 
 \cite{xu2021self} \textit{cross-view masking} augmentation to simulate occlusion hallucination in  
 multi-view situations by randomly generating a binary crop mask to block out some regions on reference
 view. The mask is then projected to other views to mask out corresponding areas in source views. They 
 also used \textit{gamma correction} to adjust the illuminance of images, \textit{random color jitter}, 
 \textit{random blur} and additive \textit{random noise} in the input images. All these data-augmentation 
 methods do not affect camera parameters.

%%%%%%%%%%%%%%%%%%%%%%%%%%%%%%%%%%%%%%%%%%%%%%%%%%%%%%%%%%%%%%%%%%%%%%%%%%%%%%%%%%%%%%%%%%%%%%%%%%%%%%%%%%%%%%%%%%%
\subsection{Semantic Information for Structural Consistency}

Humans can perform stereophotogrammetry well in ambiguous areas by exploiting more cues such as 
global perception of foreground and background, relative scaling, and semantic consistency of 
individual objects. 
Deep learning-based frameworks, operating on the \textit{color constancy hypothesis} \cite{xu2021self},
generally provide a superior performance as compared to 
traditional MVS algorithms, but both methods fail on texture-less regions, different 
lighting conditions, reflections, or \textit{color constancy ambiguity}. 
Direct application of geometric and photometric constraints in such 
regions is not helpful, but high-level semantic segmentation clues
can help. Semantic segmentation clues for a
given scene can provide abstract matching clues and 
also act as structural priors for depth estimation \cite{xu2021self}.
This also improves feature extraction in areas with cluttered backgrounds and low-texture.
In this section, we explore such depth estimation methods that include 
semantic clues in their pipeline. 

Inspired by Cheng et al. \cite{cheng2017segflow}, which incorporate semantic segmentation 
information to learn optical flow from video, Yang et al. \cite{yang2018segstereo} incorporate 
semantic feature embedding and regularize semantic cues as the loss term to 
improve disparity learning in stereo problem. Semantic feature embedding is a concatenation of 
three types of features, left-image features, left-right correlation features, and 
left-image semantic features. In addition to image and correlation features, semantic features 
provide more consistent representations of featureless regions, which helps 
solve the disparity problem. They also regularize the semantic cues loss term by warping the 
right image segmentation map to the left view and comparing it with the left 
image segmentation ground truth. Minimizing the semantic cues loss term
improves its consistency in the end-to-end learning process. Dovesi et al. \cite{dovesi2020real}
also employ semantic segmentation networks in coarse-to-fine design and utilize 
additional information in stereo matching. 

Another way of utilizing semantic information for geometric and structural consistency 
is through \textit{co-segmentation}, 
which aims to predict foreground pixels of a common object to give an 
image collection \cite{joulin2012multi}. Inspired by Casser et al. \cite{casser2019unsupervised}, 
which applied co-segmentation to learn semantic information in unsupervised monocular depth
ego-motion learning problem, Xu et al. \cite{xu2021self} apply co-segmentation on multi-view 
pairs to exploit the common semantics. They adopt non-negative matrix factorization (NMF) 
\cite{ding2005equivalence} to identify common semantic clusters among multi-view images 
during the learning process. NMF is applied to the activations of a pre-trained layer \cite{collins2018deep}
to find semantic correspondences across images. We refer to Ding et al. \cite{ding2005equivalence} for more 
details on NMF. The consistency of the co-segmentation map can be expanded across multiple views 
by warping it to other views. The semantic consistency loss is calculated as per pixel 
cross-entropy loss between the warped segmentation map $\hat{S}_i$ and the 
ground truth labels converted from the reference segmentation map $S$,s

\begin{equation}
    \mathcal{L}_{semantic} = - \sum^{N}_{i=2} \left[\frac{1}{||M_i||_1} \sum^{HW}_{j=1} f(S_j) log(\hat{s}_{i,j})M_{i,j}  \right]
\end{equation}

\noindent where $f(S_j) = one hot(arg\,max(S_j))$ and $M_i$ is the binary mask indicating valid pixels from the $i^{th}$ view
to the reference view.

%%%%%%%%%%%%%%%%%%%%%%%%%%%%%%%%%%%%%%%%%%%%%%%%%%%%%%%%%%%%%%%%%%%%%%%%%%%%%%%%%%%%%%%%%%%%%%%%%%%%%%%%%%%%%%%%%%%
\subsection{Geometric Representation Learning with Contrastive Loss}

Contrastive learning \cite{hadsell2006dimensionality} learns object representations by 
enforcing an attractive force to positive pairs and a repulsive force to negative pair 
\cite{shim2021learning}. This form of representation learning has not been explored much 
in depth estimation. Fan et al. \cite{fan2023contrastive}
use contrastive learning to pay more attention to depth distribution and 
improve the overall depth estimation process by adopting 
a non-overlapping window-based contrastive learning. Lee et al. \cite{lee2021attentive}
use contrastive learning to disentangle the camera and object motion. While these 
methods use contrastive learning for estimating depth maps, none of them use 
contrastive learning to promote the geometric representation of objects. 

\begin{figure*}[t]
\begin{center}
    \includegraphics[width=0.45\textwidth]{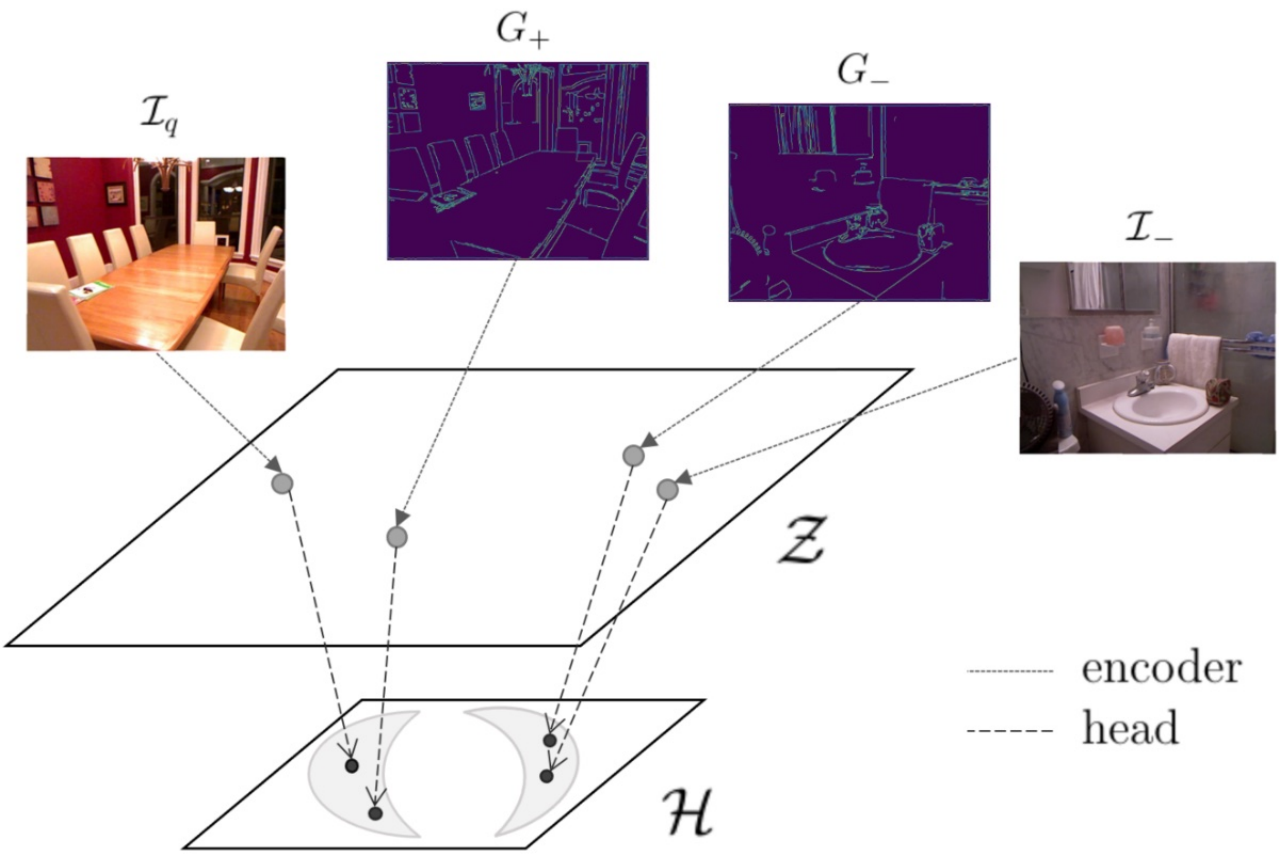}
    \caption{Contrastive learning approach to promote 
    geometric representation. Figure reprinted from \cite{shim2021learning}.}
    \label{fig:contrastive-learning}
\end{center}
\end{figure*}

Shim and Kim \cite{shim2021learning} focus on learning geometric representations 
for depth estimation using contrastive learning. They utilize 
a Canny edge binary mask \cite{canny1986computational} 
to generate gradient fields of the image as the positive 
and negative pairs, see Fig. \ref{fig:contrastive-learning}. 
To estimate the gradient field $G$ ($+$ for positive and $-$ for negative examples)
of input image pairs $\mathcal{I}_q$ (query image) and $\mathcal{I}_{-}$ (other image), 
they modify the Canny detector to extract the magnitude of the dominant gradient 
as well as its location to adjust the gradient field according to its 
edge dominance. The adopted process can be mathematically formulated as,

\begin{equation}
\begin{split}
    \mathcal{I} &\in \mathbb{R}^{h \times w}, \triangledown \mathcal{I}_x, \triangledown \mathcal{I}_y \in \mathbb{R}^{h \times w},\\
    ||E|| &= \sqrt{\mathcal{I}^2_x + \mathcal{I}^2_y},\\
    G &= B_{Canny} \odot ||E||
\end{split}
\label{eq:contrastive-gradient}
\end{equation}

\noindent where $||E||$ and $B_{Canny}$ denote the magnitude of the 
gradient from the Sobel operator and the binary mask of the Canny detector. $\odot$ is
element-wise multiplication.

The network is pre-trained with contrastive loss to learn 
the geometric representation $\mathcal{Z}$ of the images.  This 
learned representation is further compressed by a 2-layer 
fully connected network with ReLU non-linear activation to a feature 
space $\mathcal{H}$. The projected latent vector $h$ of the 
positive and the negative pairs are used to estimate the contrastive loss.

\section{Conclusion}
\label{sec:conclusion}

Progress in deep learning technologies has immensely 
benefited depth estimation by enabling the 
extraction of high-level representations from input images. 
However, it has also limited the use of modeling explicit photometric and 
geometric constraints. Most supervised stereo and MVS 
methods focus on better feature extraction and enhanced feature 
matching through attention mechanisms but 
apply a plane-sweep algorithm as the only geometric constraint. They 
largely depend on the quality of ground truth to learn about 
geometric and structural consistency in the learning process. 
In this review, we have comprehensively discussed geometric 
concepts in depth estimation and closely related domains that 
can be coupled with deep learning frameworks to enforce 
geometric and structural consistency in the learning process. 
Explicitly modeling geometric constraints, along with the supervision signal, 
can enforce structural reasoning, occlusion reasoning, better representation 
learning in regions with low-texture, cluttered background or repetitive patterns. 
It can significantly improve cross-view consistency
in any learning-based depth estimation framework. We believe this review will provide 
a good reference for readers and researchers to explore the integration of geometric 
constraints in deep learning frameworks.

% extension of the idea from Sec. \ref{subsec:geometric-consistency}

% \begin{equation}
%     \mathcal{L}_{geometric_{grad}} = \frac{1}{N} \sum^{N}_{i=1} |\Delta I_{ref} - \Delta \hat{I}^{i}_{ref \rightleftarrows src}| \odot M_{ref}
% \end{equation}

% use cost volume plus some added features to estimate source view depth maps as well. 
% put direct supervision (bidirectional supervision) section \ref{subsec:view-synthesis}

% list of data-augmentation techniques that can be applied in MVS pipeline without change in 
% camera parameters. 

%%%%%%%%%%%%%%%%%%%%%%%%%%%%%%%%% Main content END %%%%%%%%%%%%%%%%%%%%%%%%%%%%%%%%

\newpage
\bibliographystyle{unsrt}
\bibliography{references}  %%% Uncomment this line and comment out the ``thebibliography'' section below to use the external .bib file (using bibtex) .

\end{document}